\documentclass[runningheads]{llncs}
\usepackage{makeidx}
\usepackage{graphicx}
\usepackage{amsmath,amssymb} 
\usepackage{color}

\usepackage[english]{babel}
\usepackage[utf8x]{inputenc}
\usepackage{epstopdf}
\usepackage{bm}
\usepackage{multirow}
\usepackage{pdfpages}
\usepackage{xspace}
\usepackage{hyperref}
\usepackage{times}
\usepackage{epsfig}
\usepackage{float}
\usepackage{xcolor}
\usepackage{tikz-qtree}
\usepackage{tikz}
\usepackage[draft]{todonotes}   
\usepackage{algorithm}
\usepackage{mathtools,xparse}
\usepackage{siunitx}
\usepackage{stmaryrd}
\usepackage{algpseudocode}
\usepackage{placeins}
\usepackage[export]{adjustbox}
\usepackage{diagbox}
\usepackage[acronym,nomain]{glossaries}

\usepackage{amsthm}
\usepackage{dirtree}
\usepackage{algorithmicx}
\usepackage{pgfplots}
\usepackage{enumitem}
\usepackage{tkz-fct} 
\usepackage{cellspace}

\usetikzlibrary{arrows.meta}
\usetikzlibrary{trees} 
\usetikzlibrary{shapes,arrows,decorations.markings}
\usetikzlibrary{spy}
\usetikzlibrary{calc}
\usetikzlibrary{intersections}
\usetikzlibrary{matrix}

\newcommand{\C}{\mathcal{C}}

\newcommand*{\eg}{e.g.\@\xspace}
\newcommand*{\ie}{i.e.\@\xspace}
\newcommand*{\etal}{et al.\@\xspace}
\usepackage{afterpage}
\DeclarePairedDelimiter{\ceil}{\lceil}{\rceil}

\makeatletter
\def\blfootnote{\xdef\@thefnmark{}\@footnotetext}
\makeatother

\begin{document}
	\pagestyle{headings}
	\mainmatter

	\title{Non-Causal Tracking by Deblatting}

	\titlerunning{Non-Causal Tracking by Deblatting}
	\authorrunning{Denys Rozumnyi, Jan Kotera, Filip \v{S}roubek, Ji\v{r}\'{i} Matas}
\author{
Denys Rozumnyi\inst{1}\orcidID{0000-0001-9874-1349}  \and
Jan Kotera\inst{2}\orcidID{0000-0001-5528-5531}  \and \\
Filip \v{S}roubek\inst{2}\orcidID{0000-0001-6835-4911} \and
Ji\v{r}\'{i} Matas\inst{1}\orcidID{0000-0003-0863-4844}
}

\institute{
$^1$ Centre for Machine Perception, Department of Cybernetics, Faculty of Electrical Engineering, Czech Technical University in Prague, Czech Republic \\
$^2$ Institute of Information Theory and Automation, Czech Academy of Sciences, Prague, Czech Republic
}

	\maketitle

\begin{abstract}
Tracking by Deblatting\footnote{Deblatting = \textit{debl}urring and m\textit{atting}} stands for solving an inverse problem of deblurring and image matting for tracking motion-blurred objects. We propose non-causal Tracking by Deblatting which estimates continuous, complete and accurate object trajectories. Energy minimization by dynamic programming is used to detect abrupt changes of motion, called bounces. High-order polynomials are fitted to segments, which are parts of the trajectory separated by bounces. The output is a continuous trajectory function which assigns location for every real-valued time stamp from zero to the number of frames. Additionally, we show that from the trajectory function precise physical calculations are possible, such as radius, gravity or sub-frame object velocity. Velocity estimation is compared to the high-speed camera measurements and radars. Results show high performance of the proposed method in terms of Trajectory-IoU, recall and velocity estimation.
\end{abstract}

\section{Introduction}
\label{sec:introduction}

The field of visual object tracking has received huge attention in recent years~\cite{otb,vot2016,vot2018}. 
The developed techniques cover many problems and various methods were proposed, such as single object tracking~\cite{csrdcf,dsst,asms,Tang_2018_CVPR}, long-term tracking~\cite{fucolot}, methods with re-detection and learning~\cite{kalal2012tracking,uav_benchmark_eccv2016,moudgil2017long,tao2017tracking}, or multi-view~\cite{multiview} and multi-camera~\cite{multicamera} methods.

Detection and tracking of fast moving objects is an underexplored area of tracking. In a paper focusing on tracking objects that move very fast with respect to the camera, Rozumnyi~\etal~\cite{fmo} presented the first algorithm that tracks such objects,~\ie objects that satisfy the Fast Moving Object (FMO) assumption -- the object travels a distance larger than its size during exposure time. The method~\cite{fmo} operates under restrictive conditions~--~the motion-blurred object should be visible in the difference image and trajectories in each frame should be approximately linear.

Recently, a method called Tracking by Deblatting\footnotemark[\value{footnote}] (TbD) has been introduced by Kotera~\etal~\cite{tbd} to alleviate some of these restrictions. TbD performs significantly better than~\cite{fmo} and for a larger range of scenarios. The method solves two inverse problems of deblurring and image matting, and estimates object trajectories as piece-wise parabolic curves in each frame individually. 

In its core, TbD assumes causal processing of video frames, \ie the trajectory reported in the current frame is estimated using only information from previous frames. Applications of detection and tracking of fast moving objects do not usually require online and causal processing. 
Moreover, non-causal trajectory estimation brings many advantages, such as complete and accurate trajectories, which were among TbD limitations,~\eg failures at contact with a player or missing detection.

\begin{figure}[t]
\centering

\begin{tabular}{@{}c@{}c@{}c@{}}
\resizebox {!}{0.208\textwidth} {\resizebox {!}{0.2\textwidth} {\begin{tikzpicture} 
\begin{axis}[y dir=reverse, 
 xmin=1,xmax=1920, 
 ymin=1,ymax=1080, 
 xticklabels = \empty, yticklabels = \empty, 
 grid=none, axis equal image] 
\addplot graphics[xmin=1,xmax=1920,ymin=1,ymax=1080] {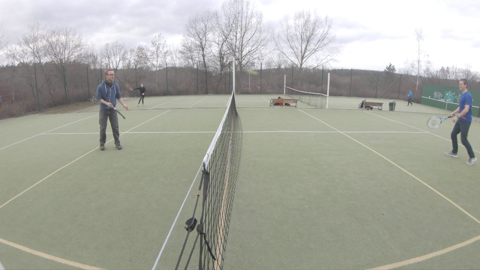}; 
\addplot [<-,>={Latex[length=1.1401mm,width=0.5mm,angle'=25,open,round]},,domain=0:1,samples=2,style=semithick,color={rgb,255:red,50; green,255; blue,0}]({1554.3065 + 17.8483*x},{354.6145 + 14.1907*x});  
\addplot [<-,>={Latex[length=1.4008mm,width=0.5mm,angle'=25,open,round]},,domain=0:1,samples=2,style=semithick,color={rgb,255:red,52; green,255; blue,0}]({1530.0803 + 23.1266*x},{337.8826 + 15.8148*x});  
\addplot [<-,>={Latex[length=1.3892mm,width=0.5mm,angle'=25,open,round]},,domain=0:1,samples=2,style=semithick,color={rgb,255:red,62; green,255; blue,0}]({1505.1614 + 24.103*x},{323.7184 + 13.8204*x});  
\addplot [<-,>={Latex[length=1.3865mm,width=0.5mm,angle'=25,open,round]},,domain=0:1,samples=2,style=semithick,color={rgb,255:red,58; green,255; blue,0}]({1478.8547 + 25.0974*x},{311.3092 + 11.7927*x});  
\addplot [<-,>={Latex[length=1.4396mm,width=0.5mm,angle'=25,open,round]},,domain=0:1,samples=2,style=semithick,color={rgb,255:red,73; green,255; blue,0}]({1451.0475 + 26.9079*x},{300.8751 + 10.242*x});  
\addplot [<-,>={Latex[length=1.4701mm,width=0.5mm,angle'=25,open,round]},,domain=0:1,samples=2,style=semithick,color={rgb,255:red,72; green,255; blue,0}]({1421.8055 + 28.2052*x},{292.6608 + 8.3029*x});  
\addplot [<-,>={Latex[length=1.4906mm,width=0.5mm,angle'=25,open,round]},,domain=0:1,samples=2,style=semithick,color={rgb,255:red,80; green,255; blue,0}]({1392.0531 + 29.2194*x},{286.7376 + 5.9162*x});  
\addplot [<-,>={Latex[length=1.5mm,width=0.5mm,angle'=25,open,round]},,domain=0:1,samples=2,style=semithick,color={rgb,255:red,67; green,255; blue,0}]({1360.7914 + 30.2887*x},{282.8037 + 3.5036*x});  
\addplot [<-,>={Latex[length=1.5mm,width=0.5mm,angle'=25,open,round]},,domain=0:1,samples=2,style=semithick,color={rgb,255:red,55; green,255; blue,0}]({1328.0022 + 32.1236*x},{280.9671 + 2.1655*x});  
\addplot [<-,>={Latex[length=1.5mm,width=0.5mm,angle'=25,open,round]},,domain=0:1,samples=10,style=semithick,color={rgb,255:red,86; green,255; blue,0}]({1293.7293 + 27.8019*x + 6.4472*x^2},{282.4861 + -5.2349*x + 4.0921*x^2});  
\addplot [<-,>={Latex[length=1.5mm,width=0.5mm,angle'=25,open,round]},,domain=0:1,samples=2,style=semithick,color={rgb,255:red,96; green,255; blue,0}]({1256.9181 + 38.0495*x},{287.321 + -4.5896*x});  
\addplot [<-,>={Latex[length=1.5mm,width=0.5mm,angle'=25,open,round]},,domain=0:1,samples=2,style=semithick,color={rgb,255:red,70; green,255; blue,0}]({1222.9258 + 36.8447*x},{293.5999 + -6.8371*x});  
\addplot [<-,>={Latex[length=1.5mm,width=0.5mm,angle'=25,open,round]},,domain=0:1,samples=10,style=semithick,color={rgb,255:red,55; green,255; blue,0}]({1187.753 + 35.3092*x + 1.0623*x^2},{303.1967 + -10.8588*x + 1.2186*x^2});  
\addplot [<-,>={Latex[length=1.5mm,width=0.5mm,angle'=25,open,round]},,domain=0:1,samples=2,style=semithick,color={rgb,255:red,71; green,255; blue,0}]({1148.8232 + 39.0161*x},{315.4674 + -12.9515*x});  
\addplot [<-,>={Latex[length=1.5mm,width=0.5mm,angle'=25,open,round]},,domain=0:1,samples=2,style=semithick,color={rgb,255:red,62; green,255; blue,0}]({1111.7637 + 38.108*x},{330.4279 + -15.7384*x});  
\addplot [->,>={Latex[length=1.5mm,width=0.5mm,angle'=25,open,round]},,domain=0:1,samples=10,style=semithick,color={rgb,255:red,82; green,255; blue,0}]({1111.2759 + -43.6949*x + 5.4095*x^2},{330.6264 + 19.4294*x + -1.0679*x^2});  
\addplot [<-,>={Latex[length=1.5mm,width=0.5mm,angle'=25,open,round]},,domain=0:1,samples=10,style=semithick,color={rgb,255:red,97; green,255; blue,0}]({1032.7201 + 33.4228*x + 7.1702*x^2},{371.1652 + -20.7935*x + -1.7532*x^2});  
\addplot [<-,>={Latex[length=1.5mm,width=0.5mm,angle'=25,open,round]},,domain=0:1,samples=2,style=semithick,color={rgb,255:red,136; green,255; blue,0}]({1001.4003 + 23.7524*x},{394.4872 + -19.5182*x});  
\addplot [<-,>={Latex[length=1.5mm,width=0.5mm,angle'=25,open,round]},,domain=0:1,samples=2,style=semithick,color={rgb,255:red,34; green,255; blue,0}]({955.9312 + 37.7614*x},{421.9014 + -26.342*x});  
\addplot [->,>={Latex[length=1.5mm,width=0.5mm,angle'=25,open,round]},,domain=0:1,samples=10,style=semithick,color={rgb,255:red,255; green,226; blue,0}]({945.0427 + 25.8306*x + -53.4991*x^2},{411.8279 + 29.1239*x + 7.8928*x^2});  
\addplot [<-,>={Latex[length=1.5mm,width=0.5mm,angle'=25,open,round]},,domain=0:1,samples=10,style=semithick,color={rgb,255:red,255; green,3; blue,0}]({830.2205 + 51.1679*x + 13.3339*x^2},{485.6358 + -17.0557*x + -8.2383*x^2});  
\addplot [<-,>={Latex[length=1.5mm,width=0.5mm,angle'=25,open,round]},,domain=0:1,samples=2,style=semithick,color=red]({786.1773 + 32.9919*x},{495.1576 + -5.0524*x});  
\addplot [<-,>={Latex[length=1.5mm,width=0.5mm,angle'=25,open,round]},,domain=0:1,samples=2,style=semithick,color={rgb,255:red,66; green,255; blue,0}]({812.9434 + 32.4653*x},{546.9435 + -32.5336*x});  
\addplot [<-,>={Latex[length=1.5mm,width=0.5mm,angle'=25,open,round]},,domain=0:1,samples=2,style=semithick,color={rgb,255:red,46; green,255; blue,0}]({780.3077 + 31.2799*x},{582.2858 + -33.669*x});  
\addplot [<-,>={Latex[length=1.5mm,width=0.5mm,angle'=25,open,round]},,domain=0:1,samples=2,style=semithick,color={rgb,255:red,54; green,255; blue,0}]({749.4065 + 29.5903*x},{618.4909 + -34.4936*x});  
\addplot [->,>={Latex[length=1.2877mm,width=0.5mm,angle'=25,open,round]},,domain=0:1,samples=2,style=semithick,color={rgb,255:red,157; green,255; blue,0}]({742.9501 + -13.5879*x},{628.6521 + -21.8771*x});  
\addplot [->,>={Latex[length=1.4841mm,width=0.5mm,angle'=25,open,round]},,domain=0:1,samples=2,style=semithick,color={rgb,255:red,100; green,255; blue,0}]({729.6199 + -16.0406*x},{605.6018 + -24.9739*x});  
\addplot [->,>={Latex[length=1.3598mm,width=0.5mm,angle'=25,open,round]},,domain=0:1,samples=2,style=semithick,color={rgb,255:red,73; green,255; blue,0}]({712.9099 + -15.48*x},{579.0623 + -22.36*x});  
\addplot [->,>={Latex[length=1.3102mm,width=0.5mm,angle'=25,open,round]},,domain=0:1,samples=2,style=semithick,color={rgb,255:red,87; green,255; blue,0}]({696.8278 + -15.7877*x},{555.13 + -20.9151*x});  
\addplot [->,>={Latex[length=1.1988mm,width=0.5mm,angle'=25,open,round]},,domain=0:1,samples=2,style=semithick,color={rgb,255:red,79; green,255; blue,0}]({681.0173 + -15.6626*x},{532.8479 + -18.1538*x});  
\addplot [<-,>={Latex[length=1.1672mm,width=0.5mm,angle'=25,open,round]},,domain=0:1,samples=2,style=semithick,color={rgb,255:red,93; green,255; blue,0}]({649.1505 + 16.2023*x},{496.8549 + 16.8049*x});  
\addplot [<-,>={Latex[length=1.0664mm,width=0.5mm,angle'=25,open,round]},,domain=0:1,samples=2,style=semithick,color={rgb,255:red,61; green,255; blue,0}]({633.9286 + 15.8164*x},{481.9736 + 14.3083*x});  
\addplot [<-,>={Latex[length=0.93864mm,width=0.5mm,angle'=25,open,round]},,domain=0:1,samples=2,style=semithick,color={rgb,255:red,73; green,255; blue,0}]({618.9831 + 14.8404*x},{469.731 + 11.4968*x});  
\addplot [<-,>={Latex[length=1.0392mm,width=0.5mm,angle'=25,open,round]},,domain=0:1,samples=2,style=semithick,color={rgb,255:red,179; green,255; blue,0}]({599.3631 + 16.3898*x},{454.5345 + 12.7823*x});  
\addplot [<-,>={Latex[length=1.0792mm,width=0.5mm,angle'=25,open,round]},,domain=0:1,samples=2,style=semithick,color={rgb,255:red,255; green,132; blue,0}]({578.4765 + 17.1028*x},{438.381 + 13.1654*x});  
\addplot [<-,>={Latex[length=1.1491mm,width=0.5mm,angle'=25,open,round]},,domain=0:1,samples=2,style=semithick,color={rgb,255:red,255; green,5; blue,0}]({556.4617 + 18.1821*x},{421.4029 + 14.0577*x});  
\end{axis} 
\end{tikzpicture} 

  \noindent}} & 

\resizebox {!}{0.208\textwidth} {\input{imgs/tbd_lt/tennis}} &

\resizebox {!}{0.208\textwidth} {\begin{tikzpicture}
\begin{axis}[samples=50,xmin=1,xmax=39,xlabel={Frame $t$},ylabel={Speed [$r / \epsilon$]},every axis x label/.style={at={(current axis.right of origin)},anchor=south east},
every axis y label/.style={at={(current axis.north west)},anchor=north west},]

\addplot[smooth,color=olive,style=thin] coordinates {
(1.25, 3.6754) (1.50, 3.3276) (1.75, 3.3276) (2.00, 3.3276) (2.25, 3.6754) (2.50, 3.0132) (2.75, 3.3276) (3.00, 3.6754) (3.25, 3.0132) (3.50, 3.6754) (3.75, 3.3935) (4.00, 3.0132) (4.25, 3.3935) (4.50, 3.3935) (4.75, 3.3935) (5.00, 3.3935) (5.25, 3.3935) (5.50, 3.1568) (5.75, 3.7940) (6.00, 3.1568) (6.25, 3.7940) (6.50, 3.1568) (6.75, 3.5839) (7.00, 3.4259) (7.25, 3.1568) (7.50, 3.5839) (7.75, 3.4259) (8.00, 3.5839) (8.25, 3.8806) (8.50, 3.4259) (8.75, 3.4259) (9.00, 3.4259) (9.25, 3.4259) (9.50, 3.7940) (9.75, 3.3276) (10.00, 3.8806) (10.25, 3.7940) (10.50, 3.3276) (10.75, 3.7647) (11.00, 3.7647) (11.25, 4.2353) (11.50, 3.3276) (11.75, 3.7647) (12.00, 3.7647) (12.25, 4.2614) (12.50, 4.2614) (12.75, 3.7647) (13.00, 4.2614) (13.25, 3.7647) (13.50, 4.2614) (13.75, 3.7940) (14.00, 4.3386) (14.25, 4.2614) (14.50, 4.3386) (14.75, 4.3386) (15.00, 4.3386) (15.25, 4.3386) (15.50, 4.3386) (15.75, 4.4644) (16.00, 4.7991) (16.25, 4.4644) (16.50, 4.4644) (16.75, 5.0684) (17.00, 4.0207) (17.25, 5.0684) (17.50, 5.0684) (17.75, 5.0684) (18.00, 4.6348) (18.25, 5.0684) (18.50, 4.8450) (18.75, 5.2613) (19.00, 5.2613) (19.25, 4.8450) (19.50, 5.2613) (19.75, 5.4880) (20.00, 5.2613) (20.25, 5.0902) (20.50, 5.4880) (20.75, 5.0902) (21.00, 5.8965) (21.25, 5.3655) (21.50, 6.1357) (21.75, 5.4880) (22.00, 6.1357) (22.25, 6.0265) (22.50, 5.7443) (22.75, 5.0902) (23.00, 6.3311) (23.25, 3.9931) (23.50, 7.3659) (23.75, 5.7443) (24.00, 5.9896) (24.25, 5.0902) (24.50, 5.6666) (24.75, 5.6666) (25.00, 4.8450) (25.25, 5.3655) (25.50, 6.0265) (25.75, 5.3241) (26.00, 5.6666) (26.25, 5.9896) (26.50, 5.3241) (26.75, 5.6666) (27.00, 5.6666) (27.25, 5.6666) (27.50, 6.0265) (27.75, 5.3655) (28.00, 6.0265) (28.25, 3.6754) (28.50, 5.2613) (28.75, 3.1568) (29.00, 4.8450) (29.25, 3.3935) (29.50, 4.0482) (29.75, 3.5839) (30.00, 3.3935) (30.25, 3.6754) (30.50, 3.3935) (30.75, 3.3935) (31.00, 3.3935) (31.25, 3.0132) (31.50, 3.7940) (31.75, 2.6620) (32.00, 3.7940) (32.25, 2.3529) (32.50, 3.0132) (32.75, 3.0132) (33.00, 3.3935) (33.25, 2.6620) (33.50, 3.0132) (33.75, 2.3529) (34.00, 3.0132) (34.25, 2.1045) (34.50, 2.6620) (34.75, 2.3529) (35.00, 2.6620) (35.25, 1.9965) (35.50, 3.0132) (35.75, 1.9403) (36.00, 2.3529) (36.25, 1.9965) (36.50, 2.1045) (36.75, 2.1045) (37.00, 2.8625) (37.25, 1.3310) (37.50, 1.4881) (37.75, 1.0523) (38.00, 1.6967) (38.25, 1.4881) (38.50, 1.4881) (38.75, 3.5839) 
} coordinate [pos=1.0] (prior) ;

\addplot[smooth,color=lightgray,style=thin] coordinates {
(3.00, 2.7662) (3.25, 2.7662) (3.50, 2.7662) (3.75, 2.7662) (4.00, 3.2481) (4.25, 3.2481) (4.50, 3.2481) (4.75, 3.2481) (5.00, 3.7734) (5.25, 3.7734) (5.50, 3.7734) (5.75, 3.7734) (6.00, 3.8390) (6.25, 3.8390) (6.50, 3.8390) (6.75, 3.8390) (7.00, 4.0190) (7.25, 4.0190) (7.50, 4.0190) (7.75, 4.0190) (8.00, 4.1943) (8.25, 4.1943) (8.50, 4.1943) (8.75, 4.1943) (9.00, 4.3068) (9.25, 4.3068) (9.50, 4.3068) (9.75, 4.3068) (10.00, 4.3898) (10.25, 4.3898) (10.50, 4.3898) (10.75, 4.3898) (11.00, 4.5807) (11.25, 4.5807) (11.50, 4.5807) (11.75, 4.5807) (12.00, 4.2923) (12.25, 4.2923) (12.50, 4.2923) (12.75, 4.2923) (13.00, 7.1985) (13.25, 7.1985) (13.50, 7.1985) (13.75, 7.1985) (14.00, 5.6592) (14.25, 5.6592) (14.50, 5.6592) (14.75, 5.6592) (15.00, 5.2851) (15.25, 5.2851) (15.50, 5.2851) (15.75, 5.2851) (16.00, 5.2030) (16.25, 5.2030) (16.50, 5.2030) (16.75, 5.2030) (17.00, 5.6170) (17.25, 5.6170) (17.50, 5.6170) (17.75, 5.6170) (18.00, 9.1658) (18.25, 9.1658) (18.50, 9.1658) (18.75, 9.1658) (19.00, 4.1810) (19.25, 4.1810) (19.50, 4.1810) (19.75, 4.1810) (20.00, 3.3958) (20.25, 3.3958) (20.50, 3.3958) (20.75, 3.3958) (21.00, 4.7352) (21.25, 4.7352) (21.50, 4.7352) (21.75, 4.7352) (22.00, 9.3697) (22.25, 9.3697) (22.50, 9.3697) (22.75, 9.3697) (23.00, 14.9146) (23.25, 14.9146) (23.50, 14.9146) (23.75, 14.9146) (24.00, 4.3229) (24.25, 4.3229) (24.50, 4.3229) (24.75, 4.3229) (25.00, 4.2552) (25.25, 4.2552) (25.50, 4.2552) (25.75, 4.2552) (26.00, 4.7336) (26.25, 4.7336) (26.50, 4.7336) (26.75, 4.7336) (27.00, 4.5883) (27.25, 4.5883) (27.50, 4.5883) (27.75, 4.5883) (28.00, 10.8188) (28.25, 10.8188) (28.50, 10.8188) (28.75, 10.8188) (29.00, 4.4889) (29.25, 4.4889) (29.50, 4.4889) (29.75, 4.4889) (30.00, 4.6005) (30.25, 4.6005) (30.50, 4.6005) (30.75, 4.6005) (31.00, 4.6061) (31.25, 4.6061) (31.50, 4.6061) (31.75, 4.6061) (32.00, 4.7288) (32.25, 4.7288) (32.50, 4.7288) (32.75, 4.7288) (33.00, 3.0419) (33.25, 3.0419) (33.50, 3.0419) (33.75, 3.0419) (34.00, 3.1809) (34.25, 3.1809) (34.50, 3.1809) (34.75, 3.1809) (35.00, 3.1204) (35.25, 3.1204) (35.50, 3.1204) (35.75, 3.1204) (36.00, 2.7247) (36.25, 2.7247) (36.50, 2.7247) (36.75, 2.7247) (37.00, 2.7939) (37.25, 2.7939) (37.50, 2.7939) (37.75, 2.7939) (38.00, 2.8855) (38.25, 2.8855) (38.50, 2.8855) (38.75, 2.8855)

} coordinate [pos=1.0] (prior) ;

\addplot [domain=1:3,color=purple]({x},{(((8.0237 + 2*(-6.393529467886)*x^1 + 3*(0.616197700909)*x^2 + 4*(-0.034127008080)*x^3 + 5*(0.000934830734)*x^4 + 6*(-0.000009739654)*x^5)^2 + (-9.4851 + 2*(-2.113442612300)*x^1 + 3*(0.295647818100)*x^2 + 4*(-0.014904339712)*x^3 + 5*(0.000411999218)*x^4 + 6*(-0.000004748125)*x^5)^2)^0.5)/8.5000}); 
\addplot [domain=3:28,color=purple]({x},{(((8.0237 + 2*(-6.393529467886)*x^1 + 3*(0.616197700909)*x^2 + 4*(-0.034127008080)*x^3 + 5*(0.000934830734)*x^4 + 6*(-0.000009739654)*x^5)^2 + (-9.4851 + 2*(-2.113442612300)*x^1 + 3*(0.295647818100)*x^2 + 4*(-0.014904339712)*x^3 + 5*(0.000411999218)*x^4 + 6*(-0.000004748125)*x^5)^2)^0.5)/8.5000}); 
\addplot [domain=28:29,color=purple]({x},{(393.7729 + -12.4234*x)/8.5000}); 
\addplot [domain=29:39,color=purple]({x},{(((-167.1665 + 2*(4.714585368625)*x^1 + 3*(-0.048880936324)*x^2)^2 + (-348.6253 + 2*(9.150031747898)*x^1 + 3*(-0.083594843498)*x^2)^2)^0.5)/8.5000});

\legend{GT, TbD, TbD-NC}
\end{axis}
\end{tikzpicture}}
\\
Tracking by Deblatting~\cite{tbd} & Non-causal trajectory estimation & Speed estimation \\
\end{tabular}

\caption[Teaser]{Trajectory reconstruction using the proposed non-causal Tracking by Deblatting (middle) compared to the causal TbD~\cite{tbd} (left). Color denotes the trajectory accuracy, from red (complete failure) to green (high accuracy). Ground truth trajectory (yellow) from high-speed camera is shown under the estimated trajectory. Speed estimation is shown on the right. Ground truth speeds (olive) are noisy due to discretization and TbD speed estimation (lightgray) is inaccurate, which is fixed by the proposed TbD-NC (purple).  }
\label{fig:teaser} 
\end{figure}

We study non-causal Tracking by Deblatting and show that global analysis of FMOs leads to accurate estimation of FMO properties, such as nearly uninterrupted trajectory, velocity and shape. 
The paper makes the following contributions:

\begin{itemize}
\item We introduce global non-causal method, referred here as TbD-NC, for estimating \textit{continuous} object trajectories by optimizing a global criterion on the whole sequence. Segments without bounces are found by an algorithm based on dynamic programming, followed by fitting of polynomials using a least squares linear program. Recovered trajectories give object location in every real-valued time stamp.
\item Compared to the causal tracker, TbD-NC reduces by a factor of 10 the number of frames where the trajectory estimation completely fails.
\item We show that TbD-NC increases the precision of the recovered trajectory to a level that allows good estimates of object velocity and size. Fig.~\ref{fig:teaser} shows an example.
\end{itemize}

\section{Related Work}
\label{sec:related}
Tracking methods that consider motion blur have been proposed in \cite{Wu2011,Seibold2017,Ma2016}, yet there is an important distinction between models therein and the problem considered here. Unlike in case of object motion, the blur is assumed to be caused by camera motion, which results in blur affecting the whole image and in the absence of alpha blending of the tracked object with the background.

To our knowledge, there are only a few published methods that tackle the problem of detection and tracking of motion-blurred objects. The first publication was the work by Rozumnyi~\etal~\cite{fmo}. The method assumes linear motion and trajectories are calculated by morphological thinning of the difference image between the given frame and the estimated background. In this paper, the first dataset with FMOs was introduced, however it contains only ground truth masks without trajectories and it cannot be used to evaluate trajectory accuracy. Deblurring of FMOs also appeared in the paper by Kotera~\etal~\cite{kotera_fmo}, focusing only on deblurring without taking into account tracking or detection.

TbD~\cite{tbd} is the only method that uses motion blur and deblurring to improve tracking results and performs parametric fit to estimate intra-frame trajectories. The TbD dataset presented therein is another dataset with FMOs which contains ground truth trajectories and can be used for evaluating trajectory accuracy. 
A brief overview of TbD follows. The acquisition model with fast moving objects proposed in~\cite{fmo,tbd} is defined as
\begin{equation}
	\label{eq:acquisition_model}
	I = H*F + (1-H*M)B,
\end{equation}
where $I$: $D\to\mathbb{R}^3$ is the current image frame defined in image domain $D \subset \mathbb{R}^2$, which is modelled by two terms. The first term is the motion-blurred object model $F$ along the trajectory given by the blur kernel $H$: $D\to\mathbb{R}$. The second term represents the influence of the background $B$ and it depends on the indicator function $M$ of object model $F$. The blur is then modelled by convolution and the background is estimated as a median of previous 3 to 5 frames. The camera is assumed to be static. We consider color images in this work and the median operator as well as convolutions are performed on each color channel separately.
TbD introduces a prior on the blur kernel $H$ and it is represented in each frame $t$ by a continuous trajectory function $\C_t$: $[0,1]\to\mathbb{R}^2$. The TbD outputs are individual trajectories $\C_t$ and blur kernels $H_t$ in every frame. 
The outputs serve as inputs to the proposed TbD-NC method.

\section{Non-Causal Tracking by Deblatting}
\label{sec:tbd}
TbD-NC is based on post-processing of individual trajectories from TbD. 
The final output of TbD-NC consists of a single trajectory $\C_f(t)$: $[0,N]\subset\mathbb{R} \to\mathbb{R}^2$, where $N$ is a number of frames in the given sequence. The function $\C_f(t)$ outputs precise object location for any real number between zero and $N$. Each frame has unit duration and the object in each frame is visible only for duration of exposure fraction $\epsilon \leq 1$. Function $\C_f(t)$ is continuous and piecewise polynomial
\begin{equation}
	\label{eq:fitting_seq_def}
	\C_f(t) = \sum_{k=0}^{d_s} \bar{c}_{s,k}t^k ~~~  t \in [t_{s-1}, t_s], s=1..S,
\end{equation}
with $S$ polynomials, where polynomial with index $s$ has degree $d_s$ and it is represented by its coefficient matrix $\bar{c}_s \in \mathbb{R}^{2,d_s}$. 
Columns of the matrix, denoted as $\bar{c}_{s,k} \in \mathbb{R}^2$, correspond to coefficients of two polynomials for $x$ and $y$ axis.
The degree depends on the size of time-frame in which the polynomial is fitted to. Variables $t_s$ form a splitting of the whole interval between 0 and $N$, \ie that $0 = t_0 < t_1 < ... < t_{S-1} < t_S = N$.

Polynomials of degree 2 (parabolic functions) can model only free falling objects under the gravitational force. In many cases forces, such as air resistance or wind, also influence the object. 
They are difficult to model mathematically by additional terms. Furthermore, we would like to keep the function linear with respect to the weights. Taylor expansion will lead to a polynomial of higher degree, which means that these forces can be approximated by adding degrees to the fitted polynomials. 
We validated experimentally that 3rd and 4th degrees are essential to explain object motion in standard scenarios. Degrees 5 and 6 provide just a small improvement, whereas degrees higher than 6 tend to overfit. Circular motion can also be approximated by~\eqref{eq:fitting_seq_def}.

A rough overview of the structure of the proposed method follows. The whole approach to estimate the piecewise polynomial function~\eqref{eq:fitting_seq_def} is based on three main steps. In the first step, the sequence is decomposed into non-intersecting parts. Each part is converted into a discrete trajectory by minimizing using dynamic programming an energy function which combines information from partial trajectories estimated by the causal TbD, curvature penalizer to force smooth trajectories and constraints on start and end points. In the second step, the discrete trajectory is further decomposed into segments by detecting bounces. Then, segments define frames which are used for fitting each polynomial. In the third step, polynomials of orders up to six are fitted into segments without bounces, which define the final trajectory function $\C_f(t)$.

\paragraph{Splitting into segments.}

When tracking fast moving objects in long-term scenarios, objects commonly move back and forth, especially in rallies. During their motion, FMOs abruptly change direction due to contact with players or when they bounce off static rigid bodies. We start with splitting the sequence into differentiable parts, \ie detecting \textit{bounces}~--~abrupt changes of object motion due to contact with other stationary or moving objects. Parts of the sequence between bounces are called \textit{segments}. Segments do not contain abrupt changes of motion and can be approximated by polynomial functions. Theoretically, causal TbD could detect bounces by fitting piecewise linear functions in one frame, but usually blur kernels are noisy and detecting bounces in just one frame is unstable. This inherent TbD instability can be fixed by non-causal processing.

To find segments and bounces, we split the sequence into \textit{non-intersecting parts} where the object does not intersect its own trajectory,~\ie either horizontal or vertical component of motion direction has the same polarity. Between non-intersecting parts we always report bounces. 
Energy minimization by dynamic programming is used to convert blur kernels $H_t$ from all frames in the given non-intersecting part into a single discrete trajectory. 
The proposed dynamic programming approach finds the global minimum of the following energy function
\begin{align}
\label{eq:dp}
\begin{split}
	E(P) = - \sum_{x=x_b}^{x_e} \sum_{t = t_{s-1}}^{t_s} H_t(x, P_x) 
	+ \kappa_1 \sum_{x=x_b+2}^{x_e} \Big| (P_x-P_{x-1}) - (P_{x-1}-P_{x-2}) \Big|   \\
	+ \kappa_2 (\C^x_{t_{s-1}}(0) - x_b) + \kappa_3 (x_e - \C^x_{t_s}(1))\,,
\end{split}
\end{align}
where variable $P$ is a discrete version of trajectory $\C$ and it is a mapping which assigns $y$ coordinate to each corresponding $x$ coordinate. $P$ is restricted to the image domain.
The first term is a data term of estimated blur kernels in all frames with the negative sign in front of the sum which accumulates more values from blur kernels while our energy function is being minimized. The second term penalizes direction changes and it is defined as the difference between directions of two following points and it is an approximation of the second order derivative of $P$. 
This term makes trajectories smoother and $\kappa_1$ serves as a smoothing parameter. The last two terms enforce that the starting point and the ending point are not far from the ones in the non-intersecting part. 
$\C^x_{t_{s-1}}(0)$ and $\C^x_{t_s}(1)$ denote $x$ coordinate of the starting point  at frame $t_{s-1}$ and the ending point at frame $t_{s}$ of causal TbD output. 
Note that in the last two terms there is no absolute value function and the sign is different, because they try to make trajectories shorter and they compete with the first term which prefers longer trajectories, \eg either making trajectory longer is worth it in terms of values in blur kernels. Without the first term, the optimal trajectory would be of zero length, \ie just a point.
Discrete trajectory $P$ is defined from $x_b$ until $x_e$ and these two variables are also being estimated. 
The ending point $\C_{t_s}(1)$ is assumed to be on the right side from the starting point $\C_{t_{s-1}}(0)$, and the image is flipped otherwise.
All $\kappa_i$ parameters were set to 0.1. 

\begin{figure}[t]
\centering
\begin{tabular}{@{}ccc@{}}
\includegraphics[frame,width=0.32\textwidth]{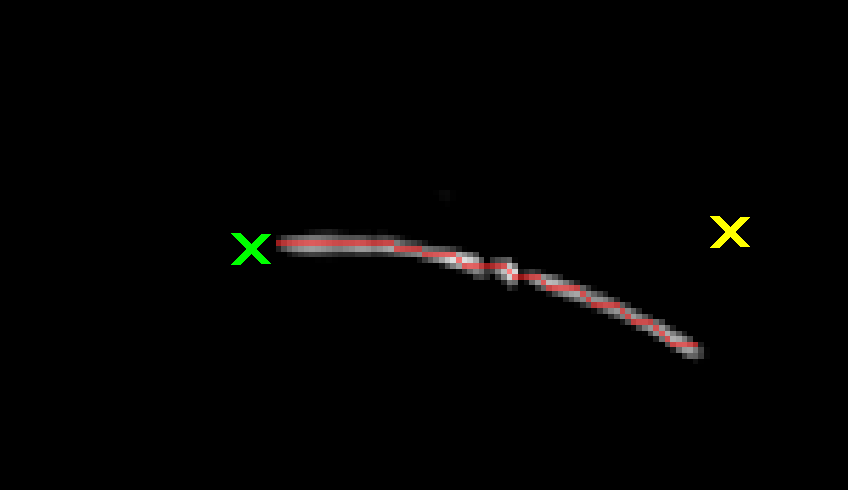} &
\includegraphics[frame,width=0.32\textwidth]{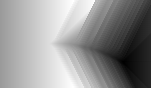} &
\includegraphics[frame,width=0.32\textwidth]{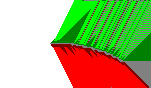} \\
\end{tabular}
\caption{Example of dynamic programming. Estimated discrete trajectory $P$ is marked in red, starting point $\C_{t_{s-1}}(0)$ by green cross, and ending point $\C_{t_s}(1)$ by yellow cross. 
These points were deliberately moved further away to show robustness of the approach. 
Left image: accumulated blur kernels from two consecutive frames $H_{t_{s-1}}$ and $H_{t_s}$ in joint coordinate system.
Middle image: value of the energy function at each pixel from black (lowest) to white (highest). 
Right image: pixels where moving down by 1 is optimal are marked in dark green, down by 2 in bright green, up by 1 in dark red, up by 2 in bright red and moving straight in grey. Pixels, where reporting a starting point $x_b$ is optimal, are white. 
The minimal value of the energy function is at the most right red pixel $x_e$ in the left image. The whole trajectory is then estimated from right to left by backtracking until the next minimizing pixel is reported as a starting point (white space). 
}
\label{fig:dp}
\end{figure}

The energy $E$~\eqref{eq:dp} is minimized by a dynamic programming (DP) approach. 
Accumulated blur kernels $H_t$ are sorted column-wise ($H_t$) or row-wise ($H_t$ transpose) to account for camera rotation or objects travelling from top to bottom. 
For both options we find the global minimum of $E$ and the one with lower energy is chosen. 
Let us illustrate the approach for the column-wise sorting. 
The row-wise case is analogous. 
DP starts with the second column and processes columns from left to right. 
We compute energy $E$ for each pixel by comparing six options and choosing the one with the lowest $E$: either adding to the trajectory one pixel out of five nearest pixels in the previous column with $y$ coordinate difference between $+2$ and $-2$, or choosing the current pixel as the starting point.
Both the minimum energy (Fig.~\ref{fig:dp} middle) and the decision option (Fig.~\ref{fig:dp} right) in every pixel is stored.
When all columns are checked, the minimum in (Fig.~\ref{fig:dp} middle) is selected as the end point and the trajectory is estimated by backtracking following decisions in (Fig.~\ref{fig:dp} right).
Backtracking finishes when a pixel is reached with the starting-point decision (white in Fig.~\ref{fig:dp} right).

\begin{figure}[t]
\centering

\begin{tabular}{@{}c@{}c@{}c@{}c@{}}

\multicolumn{2}{c}{\multirow{2}{*}{\resizebox {0.45\textwidth}{!} {\raisebox{-0.475\height}[0pt][0pt]{\includegraphics[height=0.35\textwidth]{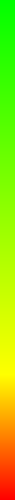} \input{imgs/segments/pingpong}}}}} &

\resizebox {0.23\textwidth}{!} {\begin{tikzpicture} 
\begin{axis}[y dir=reverse, 
 xmin=1,xmax=1280, 
 ymin=1,ymax=960, 
 xticklabels = \empty, yticklabels = \empty, 
 grid=none, axis equal image] 
\addplot graphics[xmin=1,xmax=1280,ymin=1,ymax=960] {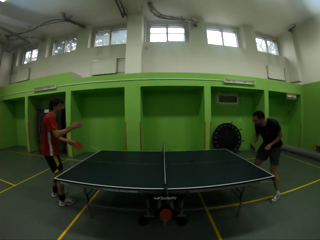}; 
\addplot [<-,>={Latex[length=1.5mm,width=0.5mm,angle'=25,open,round]},,domain=0:1,samples=2,style=semithick,color={rgb,255:red,129; green,255; blue,0}]({496.7975 + 40.1147*x},{647.4829 + -15.707*x});  
\addplot [<-,>={Latex[length=1.5mm,width=0.5mm,angle'=25,open,round]},,domain=0:1,samples=2,style=semithick,color={rgb,255:red,122; green,255; blue,0}]({455.1108 + 41.7198*x},{666.2482 + -18.6276*x});  
\addplot [<-,>={Latex[length=1.5mm,width=0.5mm,angle'=25,open,round]},,domain=0:1,samples=2,style=semithick,color={rgb,255:red,112; green,255; blue,0}]({416.9164 + 37.7805*x},{651.1807 + 17.4745*x});  
\addplot [<-,>={Latex[length=1.5mm,width=0.5mm,angle'=25,open,round]},,domain=0:1,samples=2,style=semithick,color={rgb,255:red,136; green,255; blue,0}]({382.1435 + 36.1116*x},{636.6481 + 14.7262*x});  
\addplot [<-,>={Latex[length=1.5mm,width=0.5mm,angle'=25,open,round]},,domain=0:1,samples=2,style=semithick,color={rgb,255:red,119; green,255; blue,0}]({347.1754 + 34.026*x},{624.4997 + 11.9259*x});  
\addplot [<-,>={Latex[length=1.5mm,width=0.5mm,angle'=25,open,round]},,domain=0:1,samples=2,style=semithick,color={rgb,255:red,127; green,255; blue,0}]({313.9631 + 34.0953*x},{615.1301 + 9.6637*x});  
\addplot [<-,>={Latex[length=1.5mm,width=0.5mm,angle'=25,open,round]},,domain=0:1,samples=2,style=semithick,color={rgb,255:red,255; green,119; blue,0}]({270.0412 + 35.9669*x},{602.8537 + 10.1176*x});  
\addplot [<-,>={Latex[length=1.5mm,width=0.5mm,angle'=25,open,round]},,domain=0:1,samples=2,style=semithick,color=red]({224.0479 + 37.7803*x},{589.8321 + 10.7705*x});  
\addplot [<-,>={Latex[length=1.5mm,width=0.5mm,angle'=25,open,round]},,domain=0:1,samples=2,style=semithick,color=red]({175.9741 + 39.9239*x},{576.0917 + 11.2696*x});  
\end{axis} 
\end{tikzpicture} 

  \noindent} & \resizebox {0.23\textwidth}{!} {\begin{tikzpicture} 
\begin{axis}[y dir=reverse, 
 xmin=1,xmax=1280, 
 ymin=1,ymax=960, 
 xticklabels = \empty, yticklabels = \empty, 
 grid=none, axis equal image] 
\addplot graphics[xmin=1,xmax=1280,ymin=1,ymax=960] {imgs/thumbnails/tbd/pingpong.png}; 

\addplot [->,>={Latex[length=1.5mm,width=0.5mm,angle'=25,open,round]},,domain=0:1,samples=2,style=semithick,color={rgb,255:red,125; green,255; blue,0}]({342.0063 + 46.0176*x},{598.0772 + -3.7857*x});  
\addplot [->,>={Latex[length=1.5mm,width=0.5mm,angle'=25,open,round]},,domain=0:1,samples=2,style=semithick,color={rgb,255:red,154; green,255; blue,0}]({386.9953 + 48.0017*x},{594.752 + -0.90935*x});  
\addplot [->,>={Latex[length=1.5mm,width=0.5mm,angle'=25,open,round]},,domain=0:1,samples=2,style=semithick,color={rgb,255:red,141; green,255; blue,0}]({434.0026 + 47.9909*x},{593.9071 + 1.3297*x});  
\addplot [->,>={Latex[length=1.5mm,width=0.5mm,angle'=25,open,round]},,domain=0:1,samples=2,style=semithick,color={rgb,255:red,140; green,255; blue,0}]({479.984 + 48.1161*x},{595.232 + 3.3151*x});  
\addplot [->,>={Latex[length=1.5mm,width=0.5mm,angle'=25,open,round]},,domain=0:1,samples=2,style=semithick,color={rgb,255:red,137; green,255; blue,0}]({528.0978 + 46.9666*x},{598.2649 + 6.2506*x});  
\addplot [->,>={Latex[length=1.5mm,width=0.5mm,angle'=25,open,round]},,domain=0:1,samples=2,style=semithick,color={rgb,255:red,138; green,255; blue,0}]({572.9716 + 45.1435*x},{604.181 + 7.086*x});  
\addplot [->,>={Latex[length=1.5mm,width=0.5mm,angle'=25,open,round]},,domain=0:1,samples=2,style=semithick,color={rgb,255:red,124; green,255; blue,0}]({617.1574 + 40.9839*x},{611.2888 + 9.0727*x});  
\addplot [->,>={Latex[length=1.5mm,width=0.5mm,angle'=25,open,round]},,domain=0:1,samples=2,style=semithick,color={rgb,255:red,119; green,255; blue,0}]({658.8757 + 40.3125*x},{620.4623 + 10.8377*x});  
\addplot [->,>={Latex[length=1.5mm,width=0.5mm,angle'=25,open,round]},,domain=0:1,samples=2,style=semithick,color={rgb,255:red,109; green,255; blue,0}]({698.9571 + 37.2835*x},{631.1318 + 12.1284*x});  
\addplot [->,>={Latex[length=1.5mm,width=0.5mm,angle'=25,open,round]},,domain=0:1,samples=10,style=semithick,color={rgb,255:red,127; green,255; blue,0}]({736.5982 + 46.9017*x + -14.6973*x^2},{642.9915 + 28.4135*x + -29.1449*x^2});  
\addplot [->,>={Latex[length=1.5mm,width=0.5mm,angle'=25,open,round]},,domain=0:1,samples=2,style=semithick,color={rgb,255:red,106; green,255; blue,0}]({769.7325 + 30.2261*x},{642.5125 + -16.588*x});  
\addplot [->,>={Latex[length=1.5mm,width=0.5mm,angle'=25,open,round]},,domain=0:1,samples=2,style=semithick,color={rgb,255:red,108; green,255; blue,0}]({799.9154 + 28.2828*x},{625.8214 + -13.4033*x});  
\addplot [->,>={Latex[length=1.439mm,width=0.5mm,angle'=25,open,round]},,domain=0:1,samples=2,style=semithick,color={rgb,255:red,109; green,255; blue,0}]({828.1999 + 26.691*x},{612.4956 + -10.7661*x});  
\addplot [->,>={Latex[length=1.4018mm,width=0.5mm,angle'=25,open,round]},,domain=0:1,samples=10,style=semithick,color={rgb,255:red,104; green,255; blue,0}]({854.813 + 26.8552*x + 0.020083*x^2},{601.4897 + -9.9611*x + 2.0579*x^2});  
\addplot [->,>={Latex[length=1.2528mm,width=0.5mm,angle'=25,open,round]},,domain=0:1,samples=10,style=semithick,color={rgb,255:red,98; green,255; blue,0}]({878.6346 + 23.9634*x + 0.42921*x^2},{593.8048 + -7.4406*x + 1.8193*x^2});  
\addplot [->,>={Latex[length=1.35mm,width=0.5mm,angle'=25,open,round]},,domain=0:1,samples=2,style=semithick,color={rgb,255:red,255; green,86; blue,0}]({914 + 27*x},{579.9876 + 0.010278*x});  
\addplot [->,>={Latex[length=1.45mm,width=0.5mm,angle'=25,open,round]},,domain=0:1,samples=2,style=semithick,color=red]({947 + 29*x},{580 + 8.8818e-15*x});

\end{axis} 
\end{tikzpicture} 

  \noindent} \\  & & \resizebox {0.23\textwidth}{!} {\begin{tikzpicture} 
\begin{axis}[y dir=reverse, 
 xmin=1,xmax=1280, 
 ymin=1,ymax=960, 
 xticklabels = \empty, yticklabels = \empty, 
 grid=none, axis equal image] 
\addplot graphics[xmin=1,xmax=1280,ymin=1,ymax=960] {imgs/thumbnails/tbd/pingpong.png}; 
 
\addplot [<-,>={Latex[length=1.5mm,width=0.5mm,angle'=25,open,round]},,domain=0:1,samples=2,style=semithick,color={rgb,255:red,135; green,255; blue,0}]({846.9542 + 41.011*x},{589.339 + -2.8404*x});  
\addplot [<-,>={Latex[length=1.5mm,width=0.5mm,angle'=25,open,round]},,domain=0:1,samples=2,style=semithick,color={rgb,255:red,125; green,255; blue,0}]({804.9129 + 42.0235*x},{594.236 + -4.7936*x});  
\addplot [<-,>={Latex[length=1.5mm,width=0.5mm,angle'=25,open,round]},,domain=0:1,samples=10,style=semithick,color={rgb,255:red,160; green,255; blue,0}]({759.087 + 52.859*x + -7.0333*x^2},{601.6013 + -7.4568*x + 0.34113*x^2});  
\addplot [<-,>={Latex[length=1.5mm,width=0.5mm,angle'=25,open,round]},,domain=0:1,samples=10,style=semithick,color={rgb,255:red,146; green,255; blue,0}]({709.9672 + 55.9381*x + -6.9744*x^2},{611.8437 + -13.5611*x + 4.1281*x^2});  
\addplot [<-,>={Latex[length=1.5mm,width=0.5mm,angle'=25,open,round]},,domain=0:1,samples=2,style=semithick,color={rgb,255:red,151; green,255; blue,0}]({660.0614 + 52.9304*x},{624.2448 + -13.2774*x});  
\addplot [<-,>={Latex[length=1.5mm,width=0.5mm,angle'=25,open,round]},,domain=0:1,samples=2,style=semithick,color={rgb,255:red,135; green,255; blue,0}]({611.0643 + 49.2555*x},{638.2244 + -14.108*x});  
\addplot [<-,>={Latex[length=1.5mm,width=0.5mm,angle'=25,open,round]},,domain=0:1,samples=2,style=semithick,color={rgb,255:red,128; green,255; blue,0}]({561.0098 + 50.7133*x},{654.0313 + -15.9137*x});  
\addplot [<-,>={Latex[length=1.5mm,width=0.5mm,angle'=25,open,round]},,domain=0:1,samples=2,style=semithick,color={rgb,255:red,128; green,255; blue,0}]({511.0345 + 49.9608*x},{672.0952 + -18.1083*x});  
\addplot [<-,>={Latex[length=1.5mm,width=0.5mm,angle'=25,open,round]},,domain=0:1,samples=10,style=semithick,color={rgb,255:red,227; green,255; blue,0}]({465.5085 + 71.4138*x + -26.5158*x^2},{667.8556 + 32.0163*x + -28.3343*x^2});  
\addplot [<-,>={Latex[length=1.5mm,width=0.5mm,angle'=25,open,round]},,domain=0:1,samples=2,style=semithick,color={rgb,255:red,149; green,255; blue,0}]({420.2214 + 45.707*x},{656.1641 + 12.1062*x});  
\addplot [<-,>={Latex[length=1.5mm,width=0.5mm,angle'=25,open,round]},,domain=0:1,samples=2,style=semithick,color={rgb,255:red,154; green,255; blue,0}]({377.9779 + 44.1018*x},{647.1021 + 9.5287*x});  
\addplot [<-,>={Latex[length=1.5mm,width=0.5mm,angle'=25,open,round]},,domain=0:1,samples=2,style=semithick,color={rgb,255:red,143; green,255; blue,0}]({337.1585 + 42.1081*x},{640.0962 + 7.3833*x});  
\addplot [<-,>={Latex[length=1.5mm,width=0.5mm,angle'=25,open,round]},,domain=0:1,samples=2,style=semithick,color={rgb,255:red,130; green,255; blue,0}]({298.9587 + 40.0138*x},{636.428 + 3.857*x});  
\addplot [->,>={Latex[length=1.5mm,width=0.5mm,angle'=25,open,round]},,domain=0:1,samples=10,style=semithick,color={rgb,255:red,129; green,255; blue,0}]({300.116 + -37.8733*x + 0.74457*x^2},{636.9137 + -4.1791*x + 3.0186*x^2});  
\addplot [<-,>={Latex[length=1.5mm,width=0.5mm,angle'=25,open,round]},,domain=0:1,samples=2,style=semithick,color={rgb,255:red,255; green,78; blue,0}]({217.8934 + 37.0714*x},{642.4008 + -6.5987*x});  
\addplot [<-,>={Latex[length=1.5mm,width=0.5mm,angle'=25,open,round]},,domain=0:1,samples=2,style=semithick,color=red]({170.9773 + 38.9696*x},{650.8767 + -7.1654*x});  
\addplot [<-,>={Latex[length=1.5mm,width=0.5mm,angle'=25,open,round]},,domain=0:1,samples=2,style=semithick,color=red]({120.9987 + 40.9319*x},{659.993 + -7.3779*x});  
\end{axis} 
\end{tikzpicture} 

  \noindent} & \resizebox {0.23\textwidth}{!} {\begin{tikzpicture} 
\begin{axis}[y dir=reverse, 
 xmin=1,xmax=1280, 
 ymin=1,ymax=960, 
 xticklabels = \empty, yticklabels = \empty, 
 grid=none, axis equal image] 
\addplot graphics[xmin=1,xmax=1280,ymin=1,ymax=960] {imgs/thumbnails/tbd/pingpong.png}; 

\addplot [->,>={Latex[length=1.5mm,width=0.5mm,angle'=25,open,round]},,domain=0:1,samples=2,style=semithick,color={rgb,255:red,164; green,255; blue,0}]({335.9908 + 62.0112*x},{625.7912 + -2.7461*x});  
\addplot [->,>={Latex[length=1.5mm,width=0.5mm,angle'=25,open,round]},,domain=0:1,samples=2,style=semithick,color={rgb,255:red,162; green,255; blue,0}]({396.9903 + 62.0161*x},{623.3949 + -0.99998*x});  
\addplot [->,>={Latex[length=1.5mm,width=0.5mm,angle'=25,open,round]},,domain=0:1,samples=2,style=semithick,color={rgb,255:red,150; green,255; blue,0}]({457.9902 + 61.0113*x},{622.383 + 1.5549*x});  
\addplot [->,>={Latex[length=1.5mm,width=0.5mm,angle'=25,open,round]},,domain=0:1,samples=2,style=semithick,color={rgb,255:red,145; green,255; blue,0}]({519.0114 + 60.0286*x},{623.8046 + 3.5114*x});  
\addplot [->,>={Latex[length=1.5mm,width=0.5mm,angle'=25,open,round]},,domain=0:1,samples=2,style=semithick,color={rgb,255:red,139; green,255; blue,0}]({577.9744 + 57.0958*x},{627.3003 + 4.878*x});  
\addplot [->,>={Latex[length=1.5mm,width=0.5mm,angle'=25,open,round]},,domain=0:1,samples=2,style=semithick,color={rgb,255:red,136; green,255; blue,0}]({634.9505 + 52.08*x},{632.4957 + 5.1984*x});  
\addplot [->,>={Latex[length=1.5mm,width=0.5mm,angle'=25,open,round]},,domain=0:1,samples=2,style=semithick,color={rgb,255:red,134; green,255; blue,0}]({685.978 + 49.0833*x},{637.1696 + 6.3563*x});  
\addplot [->,>={Latex[length=1.5mm,width=0.5mm,angle'=25,open,round]},,domain=0:1,samples=2,style=semithick,color={rgb,255:red,132; green,255; blue,0}]({735.0914 + 44.0113*x},{643.4935 + 7.9374*x});  
\addplot [->,>={Latex[length=1.5mm,width=0.5mm,angle'=25,open,round]},,domain=0:1,samples=10,style=semithick,color={rgb,255:red,180; green,255; blue,0}]({779.0811 + 62.6733*x + -24.9409*x^2},{650.8713 + 22.2035*x + -21.1547*x^2});  
\addplot [->,>={Latex[length=1.5mm,width=0.5mm,angle'=25,open,round]},,domain=0:1,samples=2,style=semithick,color={rgb,255:red,125; green,255; blue,0}]({817.7462 + 34.8994*x},{651.4851 + -17.204*x});  
\addplot [->,>={Latex[length=1.5mm,width=0.5mm,angle'=25,open,round]},,domain=0:1,samples=2,style=semithick,color={rgb,255:red,127; green,255; blue,0}]({852.1177 + 32.3386*x},{634.2804 + -13.5756*x});  
\addplot [->,>={Latex[length=1.5mm,width=0.5mm,angle'=25,open,round]},,domain=0:1,samples=2,style=semithick,color={rgb,255:red,132; green,255; blue,0}]({883.0485 + 30.7505*x},{621.1393 + -10.7159*x});  
\addplot [->,>={Latex[length=1.334mm,width=0.5mm,angle'=25,open,round]},,domain=0:1,samples=2,style=semithick,color={rgb,255:red,113; green,255; blue,0}]({912.8589 + 25.9027*x},{610.4285 + -6.3943*x});  
\end{axis} 
\end{tikzpicture} 

  \noindent}  \\

\resizebox {0.23\textwidth}{!} {\begin{tikzpicture} 
\begin{axis}[y dir=reverse, 
 xmin=1,xmax=1280, 
 ymin=1,ymax=960, 
 xticklabels = \empty, yticklabels = \empty, 
 grid=none, axis equal image] 
\addplot graphics[xmin=1,xmax=1280,ymin=1,ymax=960] {imgs/thumbnails/tbd/pingpong.png}; 

\addplot [>={Latex[length=1.5mm,width=0.5mm,angle'=25,open,round]},,domain=1:3,samples=20,style=semithick,color=yellow]({647.5765 + -34.4805*x + -0.80253*x^2},{601.972 + 5.5537*x + 1.4603*x^2});  

\addplot [>={Latex[length=1.5mm,width=0.5mm,angle'=25,open,round]},,domain=3:5,samples=20,style=semithick,color=green]({647.5765 + -34.4805*x + -0.80253*x^2},{601.972 + 5.5537*x + 1.4603*x^2});  
\addplot [>={Latex[length=1.5mm,width=0.5mm,angle'=25,open,round]},,domain=0:1,samples=2,style=semithick,color=magenta]({455.1108 + -2.1108*x},{666.2482 + 1.7518*x});  
\addplot [>={Latex[length=1.5mm,width=0.5mm,angle'=25,open,round]},,domain=0:1,samples=2,style=semithick,color=magenta]({453 + -34.7449*x},{668 + -16.6257*x});  
\addplot [->,>={Latex[length=1.5mm,width=0.5mm,angle'=25,open,round]},,domain=6:9,samples=30,style=semithick,color=green]({666.3802 + -45.7476*x + 0.73224*x^2},{795.6939 + -32.0345*x + 1.3302*x^2});

\end{axis} 
\end{tikzpicture} 

  \noindent} &  \resizebox {0.23\textwidth}{!} {\begin{tikzpicture} 
\begin{axis}[y dir=reverse, 
 xmin=1,xmax=1280, 
 ymin=1,ymax=960, 
 xticklabels = \empty, yticklabels = \empty, 
 grid=none, axis equal image] 
\addplot graphics[xmin=1,xmax=1280,ymin=1,ymax=960] {imgs/thumbnails/tbd/pingpong.png}; 

\addplot [>={Latex[length=1.5mm,width=0.5mm,angle'=25,open,round]},,domain=0:1,samples=2,style=semithick,color=red]({313.9631 + -33.9631*x},{615.1301 + -7.1301*x});  
\addplot [>={Latex[length=1.5mm,width=0.5mm,angle'=25,open,round]},,domain=0:1,samples=2,style=semithick,color=red]({280 + 62.0063*x},{608 + -9.9228*x});  
\addplot [>={Latex[length=1.5mm,width=0.5mm,angle'=25,open,round]},,domain=12:21,samples=90,style=semithick,color=green]({21.5061 + -4.1663*x + 3.5779*x^2 + -0.083753*x^3},{876.2039 + -45.0367*x + 2.0953*x^2 + -0.022804*x^3});  
\addplot [>={Latex[length=1.5mm,width=0.5mm,angle'=25,open,round]},,domain=0:1,samples=2,style=semithick,color=magenta]({736.2406 + 19.7594*x},{643.2602 + 6.7398*x});  
\addplot [>={Latex[length=1.5mm,width=0.5mm,angle'=25,open,round]},,domain=0:1,samples=2,style=semithick,color=magenta]({756 + 13.7325*x},{650 + -7.4875*x});  
\addplot [->,>={Latex[length=1.5mm,width=0.5mm,angle'=25,open,round]},,domain=22:28,samples=60,style=semithick,color=green]({225.0299 + 21.785*x + 0.13519*x^2},{1538.7071 + -64.5566*x + 1.0828*x^2});  

\end{axis} 
\end{tikzpicture} 

  \noindent}  &

\multicolumn{2}{c}{\multirow{2}{*}{\resizebox {0.49\textwidth}{!} {\raisebox{-0.475\height}[0pt][0pt]{\resizebox {0.45\textwidth}{!} {\begin{tikzpicture} 
\begin{axis}[y dir=reverse, 
 xmin=1,xmax=1280, 
 ymin=1,ymax=960, 
 xticklabels = \empty, yticklabels = \empty, 
 grid=none, axis equal image] 
\addplot graphics[xmin=1,xmax=1280,ymin=1,ymax=960] {imgs/thumbnails/tbd/pingpong.png}; 

\addplot [>={Latex[length=1.5mm,width=0.5mm,angle'=25,open,round]},,domain=1:3,samples=20,style=semithick,color=yellow]({647.5765 + -34.4805*x + -0.80253*x^2},{601.972 + 5.5537*x + 1.4603*x^2});  

\addplot [>={Latex[length=1.5mm,width=0.5mm,angle'=25,open,round]},,domain=3:5,samples=20,style=semithick,color=green]({647.5765 + -34.4805*x + -0.80253*x^2},{601.972 + 5.5537*x + 1.4603*x^2});  
\addplot [>={Latex[length=1.5mm,width=0.5mm,angle'=25,open,round]},,domain=0:1,samples=2,style=semithick,color=magenta]({455.1108 + -2.1108*x},{666.2482 + 1.7518*x});  
\addplot [>={Latex[length=1.5mm,width=0.5mm,angle'=25,open,round]},,domain=0:1,samples=2,style=semithick,color=magenta]({453 + -34.7449*x},{668 + -16.6257*x});  
\addplot [->,>={Latex[length=1.5mm,width=0.5mm,angle'=25,open,round]},,domain=6:9,samples=30,style=semithick,color=green]({666.3802 + -45.7476*x + 0.73224*x^2},{795.6939 + -32.0345*x + 1.3302*x^2});  

\addplot [>={Latex[length=1.5mm,width=0.5mm,angle'=25,open,round]},,domain=0:1,samples=2,style=semithick,color=red]({313.9631 + -33.9631*x},{615.1301 + -7.1301*x});  
\addplot [>={Latex[length=1.5mm,width=0.5mm,angle'=25,open,round]},,domain=0:1,samples=2,style=semithick,color=red]({280 + 62.0063*x},{608 + -9.9228*x});  
\addplot [>={Latex[length=1.5mm,width=0.5mm,angle'=25,open,round]},,domain=12:21,samples=90,style=semithick,color=green]({21.5061 + -4.1663*x + 3.5779*x^2 + -0.083753*x^3},{876.2039 + -45.0367*x + 2.0953*x^2 + -0.022804*x^3});  
\addplot [>={Latex[length=1.5mm,width=0.5mm,angle'=25,open,round]},,domain=0:1,samples=2,style=semithick,color=magenta]({736.2406 + 19.7594*x},{643.2602 + 6.7398*x});  
\addplot [>={Latex[length=1.5mm,width=0.5mm,angle'=25,open,round]},,domain=0:1,samples=2,style=semithick,color=magenta]({756 + 13.7325*x},{650 + -7.4875*x});  
\addplot [->,>={Latex[length=1.5mm,width=0.5mm,angle'=25,open,round]},,domain=22:28,samples=60,style=semithick,color=green]({225.0299 + 21.785*x + 0.13519*x^2},{1538.7071 + -64.5566*x + 1.0828*x^2});  

\addplot [>={Latex[length=1.5mm,width=0.5mm,angle'=25,open,round]},,domain=0:1,samples=2,style=semithick,color=red]({941 + -12*x},{579.9979 + 5.0021*x});  
\addplot [>={Latex[length=1.5mm,width=0.5mm,angle'=25,open,round]},,domain=0:1,samples=2,style=semithick,color=red]({929 + -41.0347*x},{585 + 1.4986*x});  
\addplot [>={Latex[length=1.5mm,width=0.5mm,angle'=25,open,round]},,domain=29:37,samples=80,style=semithick,color=green]({1502.6624 + -0.88086*x + -0.70054*x^2},{1446.953 + -61.3125*x + 1.0911*x^2});  
\addplot [>={Latex[length=1.5mm,width=0.5mm,angle'=25,open,round]},,domain=0:1,samples=2,style=semithick,color=magenta]({511.0345 + -15.0345*x},{672.0952 + 4.9048*x});  
\addplot [>={Latex[length=1.5mm,width=0.5mm,angle'=25,open,round]},,domain=0:1,samples=2,style=semithick,color=magenta]({496 + -30.0716*x},{677 + -8.7296*x});  
\addplot [->,>={Latex[length=1.5mm,width=0.5mm,angle'=25,open,round]},,domain=38:43,samples=50,style=semithick,color=green]({3678.0787 + -123.3628*x + 1.0219*x^2},{3163.3077 + -117.9358*x + 1.3757*x^2});

\addplot [>={Latex[length=1.5mm,width=0.5mm,angle'=25,open,round]},,domain=0:1,samples=2,style=semithick,color=red]({262.9873 + -26.9873*x},{635.7531 + -3.7531*x});  
\addplot [>={Latex[length=1.5mm,width=0.5mm,angle'=25,open,round]},,domain=0:1,samples=2,style=semithick,color=red]({236 + 99.9908*x},{632 + -6.2088*x});  
\addplot [>={Latex[length=1.5mm,width=0.5mm,angle'=25,open,round]},,domain=46:54,samples=80,style=semithick,color=green]({-5397.6194 + 183.6385*x + -1.2825*x^2},{2248.8223 + -68.0696*x + 0.71275*x^2});  
\addplot [>={Latex[length=1.5mm,width=0.5mm,angle'=25,open,round]},,domain=0:1,samples=2,style=semithick,color=magenta]({779.1026 + 25.8974*x},{651.4309 + 4.5691*x});  
\addplot [>={Latex[length=1.5mm,width=0.5mm,angle'=25,open,round]},,domain=0:1,samples=2,style=semithick,color=magenta]({805 + 12.7462*x},{656 + -4.5149*x});  
\addplot [->,>={Latex[length=1.5mm,width=0.5mm,angle'=25,open,round]},,domain=55:59,samples=40,style=semithick,color=green]({-5580.3689 + 196.5692*x + -1.4589*x^2},{7010.2011 + -212.3294*x + 1.7585*x^2});

\end{axis} 
\end{tikzpicture} 

  \noindent}}}}} \\

\resizebox {0.23\textwidth}{!} {\begin{tikzpicture} 
\begin{axis}[y dir=reverse, 
 xmin=1,xmax=1280, 
 ymin=1,ymax=960, 
 xticklabels = \empty, yticklabels = \empty, 
 grid=none, axis equal image] 
\addplot graphics[xmin=1,xmax=1280,ymin=1,ymax=960] {imgs/thumbnails/tbd/pingpong.png}; 

\addplot [>={Latex[length=1.5mm,width=0.5mm,angle'=25,open,round]},,domain=0:1,samples=2,style=semithick,color=red]({941 + -12*x},{579.9979 + 5.0021*x});  
\addplot [>={Latex[length=1.5mm,width=0.5mm,angle'=25,open,round]},,domain=0:1,samples=2,style=semithick,color=red]({929 + -41.0347*x},{585 + 1.4986*x});  
\addplot [>={Latex[length=1.5mm,width=0.5mm,angle'=25,open,round]},,domain=29:37,samples=80,style=semithick,color=green]({1502.6624 + -0.88086*x + -0.70054*x^2},{1446.953 + -61.3125*x + 1.0911*x^2});  
\addplot [>={Latex[length=1.5mm,width=0.5mm,angle'=25,open,round]},,domain=0:1,samples=2,style=semithick,color=magenta]({511.0345 + -15.0345*x},{672.0952 + 4.9048*x});  
\addplot [>={Latex[length=1.5mm,width=0.5mm,angle'=25,open,round]},,domain=0:1,samples=2,style=semithick,color=magenta]({496 + -30.0716*x},{677 + -8.7296*x});  
\addplot [->,>={Latex[length=1.5mm,width=0.5mm,angle'=25,open,round]},,domain=38:43,samples=50,style=semithick,color=green]({3678.0787 + -123.3628*x + 1.0219*x^2},{3163.3077 + -117.9358*x + 1.3757*x^2});  

\end{axis} 
\end{tikzpicture} 

  \noindent} & \resizebox {0.23\textwidth}{!} {\begin{tikzpicture} 
\begin{axis}[y dir=reverse, 
 xmin=1,xmax=1280, 
 ymin=1,ymax=960, 
 xticklabels = \empty, yticklabels = \empty, 
 grid=none, axis equal image] 
\addplot graphics[xmin=1,xmax=1280,ymin=1,ymax=960] {imgs/thumbnails/tbd/pingpong.png}; 

\addplot [>={Latex[length=1.5mm,width=0.5mm,angle'=25,open,round]},,domain=0:1,samples=2,style=semithick,color=red]({262.9873 + -26.9873*x},{635.7531 + -3.7531*x});  
\addplot [>={Latex[length=1.5mm,width=0.5mm,angle'=25,open,round]},,domain=0:1,samples=2,style=semithick,color=red]({236 + 99.9908*x},{632 + -6.2088*x});  
\addplot [>={Latex[length=1.5mm,width=0.5mm,angle'=25,open,round]},,domain=46:54,samples=80,style=semithick,color=green]({-5397.6194 + 183.6385*x + -1.2825*x^2},{2248.8223 + -68.0696*x + 0.71275*x^2});  
\addplot [>={Latex[length=1.5mm,width=0.5mm,angle'=25,open,round]},,domain=0:1,samples=2,style=semithick,color=magenta]({779.1026 + 25.8974*x},{651.4309 + 4.5691*x});  
\addplot [>={Latex[length=1.5mm,width=0.5mm,angle'=25,open,round]},,domain=0:1,samples=2,style=semithick,color=magenta]({805 + 12.7462*x},{656 + -4.5149*x});  
\addplot [->,>={Latex[length=1.5mm,width=0.5mm,angle'=25,open,round]},,domain=55:59,samples=40,style=semithick,color=green]({-5580.3689 + 196.5692*x + -1.4589*x^2},{7010.2011 + -212.3294*x + 1.7585*x^2});

\end{axis} 
\end{tikzpicture} 

  \noindent} & &  \\

\end{tabular}

\caption{TbD-NC processing steps. From left to right, top to bottom: causal TbD~\cite{tbd} output, splitting into segments, fitting polynomials to segments, final TbD-NC output. Top row: trajectories for all frames overlaid on the first frame, Trajectory-IoU accuracy measure color coded from red (failure) to green (success) by scale (top left corner). Bottom row: bounces between segments (magenta, red), fitted polynomials (green), extrapolation to the first and second frame (yellow). Arrows indicate motion direction. Best viewed when zoomed in a reader.}
\label{fig:segments}
\end{figure}

When each non-intersecting part is converted into 1D signal, it becomes easier to find bounces. We are looking for points with abrupt changes of direction. When $w$ pixels to the left and $w$ pixels to the right of the given point have a change of direction higher than some threshold, then this point is considered a bounce. In case of circular motion with no hard bounces, the approach finds a most suitable point to split the circle. After this step, the sequence is split into segments which are separated by bounces. 

\begin{figure}[t]
\centering
\begin{tabular}{@{}c@{}c@{}c@{}}

\resizebox {!}{0.21\textwidth} {\input{imgs/tbd/badminton_yellow}} & 
\resizebox {!}{0.21\textwidth} {\resizebox {!}{0.2\textwidth} {\begin{tikzpicture} 
\begin{axis}[y dir=reverse, 
 xmin=1,xmax=1920, 
 ymin=1,ymax=1080, 
 xticklabels = \empty, yticklabels = \empty, 
 grid=none, axis equal image] 
\addplot graphics[xmin=1,xmax=1920,ymin=1,ymax=1080] {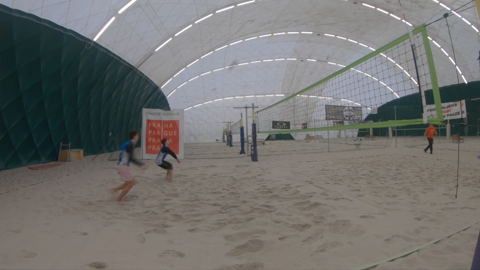}; 
\addplot [<-,>={Latex[length=1.0999mm,width=0.5mm,angle'=25,open,round]},,domain=0:1,samples=2,style=semithick,color={rgb,255:red,86; green,255; blue,0}]({862.5848 + 19.7337*x},{346.5633 + -26.4449*x});  
\addplot [<-,>={Latex[length=1.2903mm,width=0.5mm,angle'=25,open,round]},,domain=0:1,samples=2,style=semithick,color={rgb,255:red,84; green,255; blue,0}]({832.3622 + 22.2607*x},{388.4031 + -31.6682*x});  
\addplot [<-,>={Latex[length=1.3722mm,width=0.5mm,angle'=25,open,round]},,domain=0:1,samples=2,style=semithick,color={rgb,255:red,91; green,255; blue,0}]({802.8484 + 21.3277*x},{435.211 + -35.2101*x});  
\addplot [<-,>={Latex[length=1.415mm,width=0.5mm,angle'=25,open,round]},,domain=0:1,samples=2,style=semithick,color={rgb,255:red,102; green,255; blue,0}]({774.3264 + 22.0138*x},{481.698 + -36.2949*x});  
\addplot [<-,>={Latex[length=1.5mm,width=0.5mm,angle'=25,open,round]},,domain=0:1,samples=2,style=semithick,color={rgb,255:red,187; green,255; blue,0}]({746.456 + 23.3791*x},{527.6693 + -38.4656*x});  
\addplot [<-,>={Latex[length=1.5mm,width=0.5mm,angle'=25,open,round]},,domain=0:1,samples=2,style=semithick,color={rgb,255:red,234; green,255; blue,0}]({716.3048 + 25.2734*x},{577.0737 + -41.2192*x});  
\addplot [<-,>={Latex[length=1.5mm,width=0.5mm,angle'=25,open,round]},,domain=0:1,samples=2,style=semithick,color=red]({683.7009 + 26.6193*x},{629.814 + -42.8039*x});  
\addplot [->,>={Latex[length=1.1174mm,width=0.5mm,angle'=25,open,round]},,domain=0:1,samples=2,style=semithick,color={rgb,255:red,90; green,255; blue,0}]({734.659 + 4.8759*x},{494.9498 + -33.1653*x});  
\addplot [->,>={Latex[length=1.1672mm,width=0.5mm,angle'=25,open,round]},,domain=0:1,samples=2,style=semithick,color={rgb,255:red,78; green,255; blue,0}]({740.7339 + 5.7553*x},{449.9557 + -34.5407*x});  
\addplot [->,>={Latex[length=1.1672mm,width=0.5mm,angle'=25,open,round]},,domain=0:1,samples=2,style=semithick,color={rgb,255:red,53; green,255; blue,0}]({747.5407 + 5.5577*x},{406.5869 + -34.5711*x});  
\addplot [->,>={Latex[length=1.118mm,width=0.5mm,angle'=25,open,round]},,domain=0:1,samples=2,style=semithick,color={rgb,255:red,20; green,255; blue,0}]({753.525 + 5.9007*x},{366.0938 + -33.0177*x});  
\addplot [->,>={Latex[length=1.0107mm,width=0.5mm,angle'=25,open,round]},,domain=0:1,samples=2,style=semithick,color={rgb,255:red,43; green,255; blue,0}]({759.9001 + 5.3908*x},{328.3916 + -29.8391*x});  
\addplot [->,>={Latex[length=0.87251mm,width=0.5mm,angle'=25,open,round]},,domain=0:1,samples=2,style=semithick,color={rgb,255:red,79; green,255; blue,0}]({766.5554 + 4.9706*x},{294.0107 + -25.6991*x});  
\addplot [->,>={Latex[length=0.62603mm,width=0.5mm,angle'=25,open,round]},,domain=0:1,samples=2,style=semithick,color={rgb,255:red,85; green,255; blue,0}]({778.2226 + 2.2926*x},{259.8349 + -18.6405*x});  
\addplot [->,>={Latex[length=0.59762mm,width=0.5mm,angle'=25,open,round]},,domain=0:1,samples=2,style=semithick,color={rgb,255:red,65; green,255; blue,0}]({782.8767 + 4.9441*x},{232.895 + -17.2333*x});  
\addplot [->,>={Latex[length=0.58421mm,width=0.5mm,angle'=25,open,round]},,domain=0:1,samples=2,style=semithick,color={rgb,255:red,50; green,255; blue,0}]({789.7628 + 5.0053*x},{208.7273 + -16.7964*x});  
\addplot [->,>={Latex[length=0.51071mm,width=0.5mm,angle'=25,open,round]},,domain=0:1,samples=2,style=semithick,color={rgb,255:red,35; green,255; blue,0}]({795.858 + 5.6996*x},{186.3439 + -14.2215*x});  
\addplot [->,>={Latex[length=0.5mm,width=0.5mm,angle'=25,open,round]},,domain=0:1,samples=2,style=semithick,color={rgb,255:red,45; green,255; blue,0}]({803.1601 + 6.4528*x},{166.8397 + -12.5389*x});  
\addplot [->,>={Latex[length=0.5mm,width=0.5mm,angle'=25,open,round]},,domain=0:1,samples=2,style=semithick,color={rgb,255:red,44; green,255; blue,0}]({811.676 + 4.6996*x},{150.093 + -8.8945*x});  
\addplot [->,>={Latex[length=0.5mm,width=0.5mm,angle'=25,open,round]},,domain=0:1,samples=2,style=semithick,color={rgb,255:red,40; green,255; blue,0}]({818.8276 + 4.6136*x},{137.8929 + -7.4294*x});  
\addplot [->,>={Latex[length=0.5mm,width=0.5mm,angle'=25,open,round]},,domain=0:1,samples=2,style=semithick,color={rgb,255:red,71; green,255; blue,0}]({826.5 + 6.0002*x},{122.8883 + -0.00083024*x});  
\addplot [->,>={Latex[length=0.5mm,width=0.5mm,angle'=25,open,round]},,domain=0:1,samples=2,style=semithick,color={rgb,255:red,65; green,255; blue,0}]({833.9095 + 4.6623*x},{116.9835 + -0.28036*x});  
\addplot [->,>={Latex[length=0.5mm,width=0.5mm,angle'=25,open,round]},,domain=0:1,samples=2,style=semithick,color={rgb,255:red,31; green,255; blue,0}]({839.9336 + 4.6364*x},{111.1505 + -0.22819*x});  
\addplot [->,>={Latex[length=0.5mm,width=0.5mm,angle'=25,open,round]},,domain=0:1,samples=2,style=semithick,color={rgb,255:red,45; green,255; blue,0}]({847.3655 + 4.9495*x},{110.8308 + -3.9345*x});  
\addplot [<-,>={Latex[length=0.5mm,width=0.5mm,angle'=25,open,round]},,domain=0:1,samples=2,style=semithick,color={rgb,255:red,74; green,255; blue,0}]({858.3366 + -3.6606*x},{112.2786 + -5.5655*x});  
\addplot [->,>={Latex[length=0.5mm,width=0.5mm,angle'=25,open,round]},,domain=0:1,samples=2,style=semithick,color={rgb,255:red,34; green,255; blue,0}]({861.3703 + 5.4718*x},{110.6005 + 5.0698*x});  
\addplot [<-,>={Latex[length=0.5mm,width=0.5mm,angle'=25,open,round]},,domain=0:1,samples=2,style=semithick,color={rgb,255:red,18; green,255; blue,0}]({876.2207 + -6.4075*x},{122.8017 + -7.1339*x});  
\addplot [<-,>={Latex[length=0.5mm,width=0.5mm,angle'=25,open,round]},,domain=0:1,samples=2,style=semithick,color={rgb,255:red,58; green,255; blue,0}]({885.2287 + -5.6326*x},{131.8433 + -8.2241*x});  
\addplot [<-,>={Latex[length=0.5mm,width=0.5mm,angle'=25,open,round]},,domain=0:1,samples=2,style=semithick,color={rgb,255:red,52; green,255; blue,0}]({892.4407 + -6.0465*x},{145.5299 + -11.9765*x});  
\addplot [<-,>={Latex[length=0.5mm,width=0.5mm,angle'=25,open,round]},,domain=0:1,samples=2,style=semithick,color={rgb,255:red,40; green,255; blue,0}]({899.7656 + -6.5889*x},{160.6169 + -13.2062*x});  
\addplot [<-,>={Latex[length=0.58518mm,width=0.5mm,angle'=25,open,round]},,domain=0:1,samples=2,style=semithick,color={rgb,255:red,59; green,255; blue,0}]({909.1964 + -6.232*x},{179.9254 + -16.4119*x});  
\addplot [<-,>={Latex[length=0.7431mm,width=0.5mm,angle'=25,open,round]},,domain=0:1,samples=2,style=semithick,color={rgb,255:red,21; green,255; blue,0}]({917.5587 + -7.0036*x},{204.146 + -21.1643*x});  
\addplot [<-,>={Latex[length=0.79038mm,width=0.5mm,angle'=25,open,round]},,domain=0:1,samples=2,style=semithick,color={rgb,255:red,31; green,255; blue,0}]({926.3533 + -7.0022*x},{229.6999 + -22.6538*x});  
\addplot [<-,>={Latex[length=0.93356mm,width=0.5mm,angle'=25,open,round]},,domain=0:1,samples=2,style=semithick,color={rgb,255:red,41; green,255; blue,0}]({937.138 + -8.3865*x},{260.6429 + -26.7218*x});  
\addplot [<-,>={Latex[length=1.1161mm,width=0.5mm,angle'=25,open,round]},,domain=0:1,samples=2,style=semithick,color={rgb,255:red,32; green,255; blue,0}]({945.932 + -7.895*x},{295.2739 + -32.5402*x});  
\addplot [<-,>={Latex[length=1.1768mm,width=0.5mm,angle'=25,open,round]},,domain=0:1,samples=2,style=semithick,color={rgb,255:red,38; green,255; blue,0}]({955.3143 + -7.8645*x},{333.0424 + -34.4167*x});  
\addplot [<-,>={Latex[length=1.2747mm,width=0.5mm,angle'=25,open,round]},,domain=0:1,samples=2,style=semithick,color={rgb,255:red,36; green,255; blue,0}]({963.8573 + -7.2429*x},{375.124 + -37.5496*x});  
\addplot [<-,>={Latex[length=1.3238mm,width=0.5mm,angle'=25,open,round]},,domain=0:1,samples=2,style=semithick,color={rgb,255:red,40; green,255; blue,0}]({973.2991 + -7.4603*x},{420.0384 + -39.0076*x});  
\addplot [<-,>={Latex[length=1.4219mm,width=0.5mm,angle'=25,open,round]},,domain=0:1,samples=2,style=semithick,color={rgb,255:red,72; green,255; blue,0}]({983.149 + -8.253*x},{470.872 + -41.8515*x});  
\end{axis} 
\end{tikzpicture} 

  \noindent}} &  
\resizebox {!}{0.21\textwidth} {\resizebox {!}{0.2\textwidth} {\begin{tikzpicture} 
\begin{axis}[y dir=reverse, 
 xmin=1,xmax=960, 
 ymin=1,ymax=600, 
 xticklabels = \empty, yticklabels = \empty, 
 grid=none, axis equal image] 
\addplot graphics[xmin=1,xmax=960,ymin=1,ymax=600] {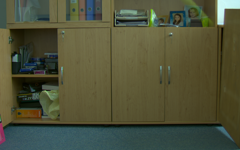}; 
\addplot [<-,>={Latex[length=1.5mm,width=0.5mm,angle'=25,open,round]},,domain=0:1,samples=2,style=semithick,color={rgb,255:red,113; green,255; blue,0}]({769.502 + 77.4291*x},{297.002 + -77.0735*x});  
\addplot [<-,>={Latex[length=1.5mm,width=0.5mm,angle'=25,open,round]},,domain=0:1,samples=2,style=semithick,color={rgb,255:red,56; green,255; blue,0}]({679.7764 + 84.6658*x},{400.9474 + -98.4272*x});  
\addplot [<-,>={Latex[length=1.5mm,width=0.5mm,angle'=25,open,round]},,domain=0:1,samples=10,style=semithick,color={rgb,255:red,73; green,255; blue,0}]({595.8696 + 91.8952*x + -11.8955*x^2},{515.8494 + -127.5539*x + 19.228*x^2});  
\addplot [->,>={Latex[length=1.5mm,width=0.5mm,angle'=25,open,round]},,domain=0:1,samples=2,style=semithick,color={rgb,255:red,84; green,255; blue,0}]({590.2227 + -70.4596*x},{515.1313 + -86.7511*x});  
\addplot [<-,>={Latex[length=1.5mm,width=0.5mm,angle'=25,open,round]},,domain=0:1,samples=10,style=semithick,color={rgb,255:red,43; green,255; blue,0}]({441.2487 + 74.0552*x + -1.0922*x^2},{352.2135 + 62.3832*x + 8.6702*x^2});  
\addplot [<-,>={Latex[length=1.5mm,width=0.5mm,angle'=25,open,round]},,domain=0:1,samples=2,style=semithick,color={rgb,255:red,61; green,255; blue,0}]({367.983 + 72.6916*x},{294.8567 + 54.5788*x});  
\addplot [<-,>={Latex[length=1.5mm,width=0.5mm,angle'=25,open,round]},,domain=0:1,samples=2,style=semithick,color={rgb,255:red,77; green,255; blue,0}]({291.4055 + 75.571*x},{254.2222 + 39.4049*x});  
\addplot [<-,>={Latex[length=1.5mm,width=0.5mm,angle'=25,open,round]},,domain=0:1,samples=2,style=semithick,color={rgb,255:red,42; green,255; blue,0}]({214.4452 + 74.0089*x},{232.7132 + 19.0199*x});  
\addplot [<-,>={Latex[length=1.5mm,width=0.5mm,angle'=25,open,round]},,domain=0:1,samples=2,style=semithick,color={rgb,255:red,80; green,255; blue,0}]({139.6245 + 77.9128*x},{227.3478 + 4.5074*x});  
\addplot [<-,>={Latex[length=1.5mm,width=0.5mm,angle'=25,open,round]},,domain=0:1,samples=10,style=semithick,color={rgb,255:red,50; green,255; blue,0}]({69.2085 + 68.6378*x + 1.6536*x^2},{242.1527 + -22.8971*x + 10.3714*x^2});  
\addplot [->,>={Latex[length=1.3665mm,width=0.5mm,angle'=25,open,round]},,domain=0:1,samples=2,style=semithick,color={rgb,255:red,129; green,255; blue,0}]({30.3938 + 36.3493*x},{261.8783 + -18.9531*x});  
\addplot [->,>={Latex[length=1.5mm,width=0.5mm,angle'=25,open,round]},,domain=0:1,samples=2,style=semithick,color={rgb,255:red,23; green,255; blue,0}]({64.305 + 51.6766*x},{255.6383 + 15.7846*x});  
\addplot [->,>={Latex[length=1.5mm,width=0.5mm,angle'=25,open,round]},,domain=0:1,samples=10,style=semithick,color={rgb,255:red,54; green,255; blue,0}]({116.7032 + 61.5336*x + -4.3684*x^2},{272.2699 + 23.1777*x + 10.706*x^2});  
\addplot [->,>={Latex[length=1.5mm,width=0.5mm,angle'=25,open,round]},,domain=0:1,samples=2,style=semithick,color={rgb,255:red,50; green,255; blue,0}]({175.109 + 55.7538*x},{304.957 + 47.7033*x});  
\addplot [<-,>={Latex[length=1.5mm,width=0.5mm,angle'=25,open,round]},,domain=0:1,samples=2,style=semithick,color={rgb,255:red,47; green,255; blue,0}]({288.4268 + -55.9352*x},{421.1362 + -65.6291*x});  
\addplot [->,>={Latex[length=1.5mm,width=0.5mm,angle'=25,open,round]},,domain=0:1,samples=10,style=semithick,color={rgb,255:red,116; green,255; blue,0}]({289.6456 + 43.0125*x + 15.485*x^2},{425.8723 + 51.2815*x + 33.6897*x^2});  
\addplot [->,>={Latex[length=1.5mm,width=0.5mm,angle'=25,open,round]},,domain=0:1,samples=2,style=semithick,color={rgb,255:red,56; green,255; blue,0}]({354.745 + 23.3464*x},{518.5542 + -49.7157*x});  
\addplot [->,>={Latex[length=1.5mm,width=0.5mm,angle'=25,open,round]},,domain=0:1,samples=2,style=semithick,color={rgb,255:red,32; green,255; blue,0}]({377.3389 + 27.0497*x},{465.5113 + -44.3839*x});  
\addplot [->,>={Latex[length=1.2736mm,width=0.5mm,angle'=25,open,round]},,domain=0:1,samples=2,style=semithick,color={rgb,255:red,24; green,255; blue,0}]({404.4616 + 27.3837*x},{417.9467 + -26.6471*x});  
\addplot [->,>={Latex[length=1.0497mm,width=0.5mm,angle'=25,open,round]},,domain=0:1,samples=2,style=semithick,color={rgb,255:red,32; green,255; blue,0}]({431.4289 + 29.1387*x},{390.1063 + -11.9414*x});  
\addplot [->,>={Latex[length=0.99413mm,width=0.5mm,angle'=25,open,round]},,domain=0:1,samples=10,style=semithick,color={rgb,255:red,25; green,255; blue,0}]({461.3835 + 29.1146*x + -1.4769*x^2},{380.4724 + -10.4826*x + 16.5526*x^2});  
\addplot [->,>={Latex[length=1.1711mm,width=0.5mm,angle'=25,open,round]},,domain=0:1,samples=2,style=semithick,color={rgb,255:red,28; green,255; blue,0}]({491.0179 + 27.5985*x},{383.3426 + 21.7402*x});  
\addplot [->,>={Latex[length=1.5mm,width=0.5mm,angle'=25,open,round]},,domain=0:1,samples=10,style=semithick,color={rgb,255:red,36; green,255; blue,0}]({517.0835 + 31.8111*x + -3.4831*x^2},{406.938 + 31.9636*x + 6.9215*x^2});  
\addplot [<-,>={Latex[length=1.5mm,width=0.5mm,angle'=25,open,round]},,domain=0:1,samples=2,style=semithick,color={rgb,255:red,43; green,255; blue,0}]({574.4917 + -27.7115*x},{504.7703 + -56.6519*x});  
\addplot [->,>={Latex[length=1.2064mm,width=0.5mm,angle'=25,open,round]},,domain=0:1,samples=2,style=semithick,color={rgb,255:red,118; green,255; blue,0}]({581.7972 + 20.0682*x},{528.8643 + -30.1202*x});  
\addplot [->,>={Latex[length=1.4603mm,width=0.5mm,angle'=25,open,round]},,domain=0:1,samples=2,style=semithick,color={rgb,255:red,40; green,255; blue,0}]({602.2376 + 26.7274*x},{495.9129 + -34.7099*x});  
\addplot [->,>={Latex[length=1.0945mm,width=0.5mm,angle'=25,open,round]},,domain=0:1,samples=10,style=semithick,color={rgb,255:red,24; green,255; blue,0}]({631.6198 + 21.3767*x + 5.4443*x^2},{459.0904 + -25.3496*x + 7.1243*x^2});  
\addplot [<-,>={Latex[length=0.90653mm,width=0.5mm,angle'=25,open,round]},,domain=0:1,samples=10,style=semithick,color={rgb,255:red,38; green,255; blue,0}]({687.3445 + -27.5626*x + 1.6255*x^2},{439.8926 + -11.2184*x + 13.5934*x^2});  
\addplot [->,>={Latex[length=1.0408mm,width=0.5mm,angle'=25,open,round]},,domain=0:1,samples=10,style=semithick,color={rgb,255:red,30; green,255; blue,0}]({688.4758 + 30.1372*x + -2.8747*x^2},{439.7186 + 6.5343*x + 7.9114*x^2});  
\addplot [<-,>={Latex[length=1.4448mm,width=0.5mm,angle'=25,open,round]},,domain=0:1,samples=2,style=semithick,color={rgb,255:red,34; green,255; blue,0}]({743.1772 + -27.1326*x},{488.1604 + -33.8017*x});  
\addplot [->,>={Latex[length=1.5mm,width=0.5mm,angle'=25,open,round]},,domain=0:1,samples=10,style=semithick,color={rgb,255:red,44; green,255; blue,0}]({742.1305 + 27.3783*x + -2.646*x^2},{488.2132 + 39.216*x + 2.6365*x^2});  
\addplot [->,>={Latex[length=1.4708mm,width=0.5mm,angle'=25,open,round]},,domain=0:1,samples=10,style=semithick,color={rgb,255:red,38; green,255; blue,0}]({768.8735 + 28.3971*x + 0.49244*x^2},{523.0727 + -42.2063*x + 9.055*x^2});  
\addplot [->,>={Latex[length=1.0547mm,width=0.5mm,angle'=25,open,round]},,domain=0:1,samples=2,style=semithick,color={rgb,255:red,16; green,255; blue,0}]({797.7462 + 26.9557*x},{488.5872 + -16.572*x});  
\addplot [->,>={Latex[length=0.90009mm,width=0.5mm,angle'=25,open,round]},,domain=0:1,samples=2,style=semithick,color={rgb,255:red,38; green,255; blue,0}]({826.5003 + 27.0028*x},{472.3494 + 0.05228*x});  
\addplot [->,>={Latex[length=1.0404mm,width=0.5mm,angle'=25,open,round]},,domain=0:1,samples=2,style=semithick,color={rgb,255:red,42; green,255; blue,0}]({853.9395 + 25.5624*x},{474.8727 + 17.9109*x});  
\addplot [->,>={Latex[length=0.77977mm,width=0.5mm,angle'=25,open,round]},,domain=0:1,samples=2,style=semithick,color={rgb,255:red,106; green,255; blue,0}]({893.4649 + -13.9354*x},{510.7842 + -18.7894*x});  
\addplot [->,>={Latex[length=0.56997mm,width=0.5mm,angle'=25,open,round]},,domain=0:1,samples=2,style=semithick,color={rgb,255:red,63; green,255; blue,0}]({874.0218 + -4.1715*x},{529.2429 + -16.5826*x});  
\addplot [<-,>={Latex[length=0.5mm,width=0.5mm,angle'=25,open,round]},,domain=0:1,samples=2,style=semithick,color={rgb,255:red,31; green,255; blue,0}]({853.2797 + 12.7081*x},{509.4458 + 2.9601*x});  
\addplot [<-,>={Latex[length=0.70748mm,width=0.5mm,angle'=25,open,round]},,domain=0:1,samples=2,style=semithick,color={rgb,255:red,34; green,255; blue,0}]({835.3605 + 14.9261*x},{526.3783 + -15.0893*x});  
\addplot [<-,>={Latex[length=0.65098mm,width=0.5mm,angle'=25,open,round]},,domain=0:1,samples=10,style=semithick,color={rgb,255:red,31; green,255; blue,0}]({822.5105 + 17.0374*x + -1.4557*x^2},{516.6798 + 9.6798*x + 2.0065*x^2});  
\addplot [<-,>={Latex[length=0.63472mm,width=0.5mm,angle'=25,open,round]},,domain=0:1,samples=10,style=semithick,color={rgb,255:red,29; green,255; blue,0}]({805.1919 + 14.5981*x + 1.6977*x^2},{526.014 + -12.3616*x + 2.6822*x^2});  
\addplot [<-,>={Latex[length=0.55862mm,width=0.5mm,angle'=25,open,round]},,domain=0:1,samples=2,style=semithick,color={rgb,255:red,35; green,255; blue,0}]({790.5595 + 14.6975*x},{521.8913 + 8.0521*x});  
\addplot [<-,>={Latex[length=0.57989mm,width=0.5mm,angle'=25,open,round]},,domain=0:1,samples=2,style=semithick,color={rgb,255:red,28; green,255; blue,0}]({774.9455 + 15.3985*x},{529.7983 + -8.0951*x});  
\end{axis} 
\end{tikzpicture} 

  \noindent}} \\


\resizebox {!}{0.21\textwidth} {\input{imgs/tbd_lt/badminton_yellow}} & 
\resizebox {!}{0.21\textwidth} {\input{imgs/tbd_lt/volleyball}} & 
\resizebox {!}{0.21\textwidth} {\input{imgs/tbd_lt/throw_tennis}}  \\

\end{tabular}

\caption{Trajectory recovery for selected sequences from the TbD dataset. Top row: trajectories estimated by the causal TbD~\cite{tbd} overlaid on the first frame. TIoU~\eqref{eq:exp_TIoU_def} with ground truth trajectories from a high-speed camera is color coded by scale in Fig.~\ref{fig:segments}. Bottom row: trajectory estimates by the proposed TbD-NC which outputs continuous trajectory for the whole sequence. The yellow curves underneath denote ground truth. 
Arrows indicate the direction of motion.}
\label{fig:tbd_lt_imgs}
\end{figure}

\paragraph{Fitting polynomials.}
The output discrete trajectory $P$ has a two-fold purpose. First, it is used to estimate bounces and define segments, and second to estimate which frames belong to the segment and should be considered for fitting polynomials. To this end, we assign starting and ending points of each frame,~\ie $\C_t(0)$ and $\C_t(1)$, to the closest segment. For fitting we use only frames that completely belong to the segment, \ie $\C_t(0)$ and $\C_t(1)$ are closer to this segment than to any other.
The degree of a polynomial is a function of the number of frames ($N_s = t_s - t_{s-1} + 1$) belonging to the segment 
\begin{equation}
	\label{eq:degree}	
	d_s = \min(6, \ceil{N_s/3} ).
\end{equation}
The polynomial coefficients are found by solving a linear least-squares problem


\begin{equation} \label{eq:lsqlin}	
\begin{array}{l@{}l}
\min_{\bar{c}_{s}} ~~ & \sum_{t=t_{s-1}}^{t_s}  \| \C_f(t)  - \C_t(0) \|^2 + \| \C_f(t+\epsilon)  - \C_t(1) \|^2 \\
~~ \text{s. t.} ~~~ & \C_f(t_{s-1}) = \C_{t_{s-1}}(0) ~~ ~~ \text{and} ~~  ~~ \C_f(t_s+\epsilon) = \C_{t_s}(1), \\
\end{array}
\end{equation}
where $s$ denotes the segment index. Equality constraints force continuity of the curve throughout the whole sequence,~\ie we get curves of differentiability class $C^0$. The least-squares objective enforces similarity to the trajectories estimated during the causal TbD pipeline. The final trajectory $\C_f$ is defined over the whole sequence and the last visible point in the frame $t$ which is $\C_t(1)$ corresponds to $\C_f(t+\epsilon)$ in the sequence time-frame, where the exposure fraction $\epsilon$ is assumed to be constant in the sequence. The exposure fraction is estimated as an average ratio of the length of trajectories $\C_t$ in each frame and the distance between adjacent starting points
\begin{equation} \label{eq:exp_est}	
\epsilon = \frac{1}{N-1} \sum_{t=1}^{N-1} \frac{\|\C_t(1) - \C_t(0)\|}{\|\C_{t+1}(0) - \C_t(0)\|}.
\end{equation}

Frames which are only partially in segments contain bounces. We replace them with a piecewise linear polynomial which connects the last point from the previous segment, bounce point found by dynamic programming and the first point from the following segment. Frames between non-intersecting parts are also interpolated by piecewise linear polynomial which connects the last point of the previous segment, point of intersection of these two segments and the first point of the following segment. Frames which are before the first detection or after the last non-empty $\C_t$ are extrapolated by the closest segment. Fig.~\ref{fig:segments} shows an example of splitting a sequence into segments which are used for fitting polynomials. More examples of full trajectory estimation are in Fig.~\ref{fig:tbd_lt_imgs}.

\section{Experiments}
\label{sec:eval}
\begin{table}[t]
\centering
\setlength{\tabcolsep}{1.8mm}
\caption[Performance on TbD dataset]{TIoU~\eqref{eq:exp_TIoU_def} and recall (Rcl) on the TbD dataset -- comparison of TbD, FuCoLoT, FMO methods and the proposed TbD-NC. FuCoLoT is a standard, well-performing~\cite{vot2018}, near real-time tracker. For each sequence, the highest TIoU is highlighted in blue and recall in cyan.}
\begin{center}
\begin{tabular}{l|c|c|c|c|c|c|c|c|c}
\hline
\multirow{2}{*}{Sequence} & \multirow{2}{*}{Frames} & \multicolumn{2}{c|}{FuCoLoT~\cite{fucolot}}   & \multicolumn{2}{c|}{FMO~\cite{fmo}}  & \multicolumn{2}{c|}{TbD~\cite{tbd}} & \multicolumn{2}{c}{TbD-NC} \\ 
 \cline{3-10}
 & & TIoU & Rcl & TIoU & Rcl & TIoU & Rcl & TIoU & Rcl  \\ \hline

badminton\_white & 40 & .286 & 0.39 & .242 & 0.34 & .694 & 0.97 & \textcolor{blue}{.783} & \textcolor{cyan}{1.00}\\  
 badminton\_yellow & 57 & .123 & 0.22 & .236 & 0.31 & .677 & 0.91 & \textcolor{blue}{.780} & \textcolor{cyan}{1.00}\\  
 pingpong & 58 & .065 & 0.14 & .064 & 0.12 & .523 & 0.91 & \textcolor{blue}{.643} & \textcolor{cyan}{1.00}\\  
 tennis & 38 & .294 & 0.89 & .596 & 0.78 & .673 & 0.97 & \textcolor{blue}{.750} & \textcolor{cyan}{1.00}\\  
 volleyball & 41 & .496 & 0.79 & .537 & 0.72 & .795 & 0.97 & \textcolor{blue}{.857} & \textcolor{cyan}{1.00}\\  
 throw\_floor & 40 & .275 & 0.63 & .272 & 0.37 & .810 & \textcolor{cyan}{1.00} & \textcolor{blue}{.855} & \textcolor{cyan}{1.00}\\  
 throw\_soft & 60 & .463 & 0.95 & .377 & 0.57 & .652 & 0.97 & \textcolor{blue}{.761} & \textcolor{cyan}{1.00}\\  
 throw\_tennis & 45 & .239 & 0.98 & .507 & 0.65 & .850 & \textcolor{cyan}{1.00} & \textcolor{blue}{.878} & \textcolor{cyan}{1.00}\\  
 roll\_golf & 16 & .360 & \textcolor{cyan}{1.00} & .187 & 0.71 & .873 & \textcolor{cyan}{1.00} & \textcolor{blue}{.894} & \textcolor{cyan}{1.00}\\  
 fall\_cube & 20 & .324 & 0.67 & .408 & 0.78 & .721 & \textcolor{cyan}{1.00} & \textcolor{blue}{.757} & \textcolor{cyan}{1.00}\\  
 hit\_tennis & 30 & .330 & 0.93 & .381 & 0.68 & .667 & 0.93 & \textcolor{blue}{.714} & \textcolor{cyan}{1.00}\\  
 hit\_tennis2 & 26 & .226 & 0.79 & .414 & 0.71 & .616 & 0.83 & \textcolor{blue}{.682} & \textcolor{cyan}{0.92}\\  
 \hline 
  Average  & 39 & .290 & 0.70 & .352 & 0.56 & .713 & 0.96 & \textcolor{blue}{.779} & \textcolor{cyan}{0.99}\\

\hline
\end{tabular}
\end{center}
\label{tbl:sota}
\end{table}

Experiments are done on the TbD dataset~\cite{tbd} with the ground truth trajectories from a high-speed camera. 
We use Trajectory Intersection over Union (TIoU) proposed by Kotera~\etal~\cite{tbd} to measure the accuracy of estimated trajectories, which is defined as
\begin{equation}
	\label{eq:exp_TIoU_def}
	\operatorname{TIoU}(\mathcal{C},\mathcal{C}^*) = \int_{t} \operatorname{IoU}\left(\rule{0pt}{2ex}M^*_{\mathcal{C}(t)},\, M^*_{\mathcal{C}^*(t)}\right)\mathrm{d}t,
\end{equation}
where the estimated trajectory $\C$ is compared to the ground-truth trajectory $\C^*$. 
The ground truth object appearance mask $M^*$ is used to measure IoU at different points $x$ on the trajectory, denoted by $M^*_x$. Time $t$ is discretized into 10 evenly spaced time-stamps to approximate integral.

\begin{table} [t]
\centering
\setlength{\tabcolsep}{3.8mm}
\caption[TbD Failure]{Comparison of TbD-NC with TbD~\cite{tbd}. TbD failure is defined as frames where TIoU~\eqref{eq:exp_TIoU_def} equals to zero. TbD-NC decreases the number of frames with failure by a factor of 10.}
\begin{center}
\begin{tabular}{c|c|c|c|c}
\hline
  &   TbD [TIoU]  & TbD-NC [TIoU] &  TbD [\%] & TbD-NC [\%] \\
\hline
 TbD  Fails     &  0.000  & 0.382  & 4.7 & 0.4  \\
 TbD TIoU$ > 0$  &  0.744  & 0.800  & 95.3   & 99.6  \\
\hline
\end{tabular}
\end{center}
\label{tbl:tbd_fail}
\end{table}

Comparison to baselines on the TbD dataset is presented in Table~\ref{tbl:sota}. We use the recently introduced long-term tracker FuCoLoT~\cite{fucolot} as a baseline standard tracker, the FMO method~\cite{fmo} as a baseline for a tracker specialized on fast moving objects and Tracking by Deblatting~\cite{tbd} (causal TbD with a template) as a well-performing method for establishing trajectories in each frame. The proposed TbD-NC outperforms all baselines in both recall and TIoU. 
Recall is 100\% in all cases except one, where the first detection appeared only on the seventh frame and extrapolation to the first six frames was not successful. 
Table~\ref{tbl:tbd_fail} shows that TbD-NC corrects complete failures of causal TbD when TIoU is zero, \eg due to wrong predictions or other moving objects. TbD-NC also improves TIoU of successful detection by fixing small local errors, \eg when the blur is misleading or fitting in one frame is not precise.

\input{figures/speed_estimation_small}

\paragraph{Speed estimation.}
Tbd-NC provides the trajectory function $\C_f(t)$, which is defined for each real-valued time stamp $t$ between 0 and the number of frames. Taking the norm of the derivative of $\C_f(t)$ gives a real-valued function of object velocity, measured in pixels per exposure. To normalize it with respect to the object, we divide it by the radius and report speed in radii per exposure. The results are visualized in Fig.~\ref{fig:speed} where sequences are shown together with their speed functions. The ground-truth speed was estimated from a high-speed camera footage having 8 times higher frame rate. The object center was detected in every frame and the GT speed was then calculated from the distance between the object centers in adjacent frames. 
Deliberately, we used no prior information (regularization) to smooth the GT speed and therefore it is noisy as can be seen in Fig.~\ref{fig:speed}.
We also report average absolute differences between GT and the estimated speed in Table~\ref{tbl:est}. The error is mostly due to the noise in GT. 

\begin{table}[t]
\centering
\setlength{\tabcolsep}{2.5mm}
\caption{Speed estimation compared to the radar gun (GT). We used the last 10 serves of the final match of 2010 ATP World Tour. 
The lowest error for each serve is marked in blue. }
\begin{center}
\begin{tabular}{c|c|c|c|c|c|c}
\hline
\multirow{2}{*}{Serve} & Duration & GT & \multicolumn{2}{c|}{Hrabalík~\cite{hrabalik}} & \multicolumn{2}{c}{TbD-NC}  \\ \cline{4-7}
 & [frames] & [mph]  & Speed [mph] & Error [\%] & Speed [mph] & Error [\%] \\

\hline

1 & 23 & 108 & 105.6 & 2.2 & 108.0 & \textcolor{blue}{0.0}\\  
2 & 32 & 101 & 103.8 & 2.8 & 101.6 & \textcolor{blue}{0.6}\\  
3 & 62 & 104 & 106.5 & \textcolor{blue}{2.4} & 110.4 & 6.1\\  
4 & 75 & 113 & 101.7 & 10.0 & 115.8 & \textcolor{blue}{2.5}\\  
5 & 82 & 104 & 91.9 & 11.6 & 106.9 & \textcolor{blue}{2.8}\\  
6 & 30 & 127 & 127.4 & \textcolor{blue}{0.3} & 126.3 & 0.6\\  
7 & 34 & 112 & 116.1 & \textcolor{blue}{3.7} & 107.5 & 4.0\\  
8 & 78 & 125 & 123.2 & \textcolor{blue}{1.4} & 130.3 & 4.2\\  
9 & 67 & 99 & 88.3 & 10.8 & 89.7 & \textcolor{blue}{9.4}\\  
10 & 90 & 108 & 110.2 & 2.0 & 106.2 & \textcolor{blue}{1.6}\\  
\hline 
Mean  & 57 & 110.1 & 107.5 & 4.7 & 110.3 & \textcolor{blue}{3.2}\\

\hline
\end{tabular}
\end{center}
\label{tbl:atp}
\end{table}

\begin{figure}[t]
\centering
\begin{tabular}{@{}c@{\hspace{0.3em}}c@{\hspace{0.3em}}c@{}}
\includegraphics[height=0.2\textwidth]{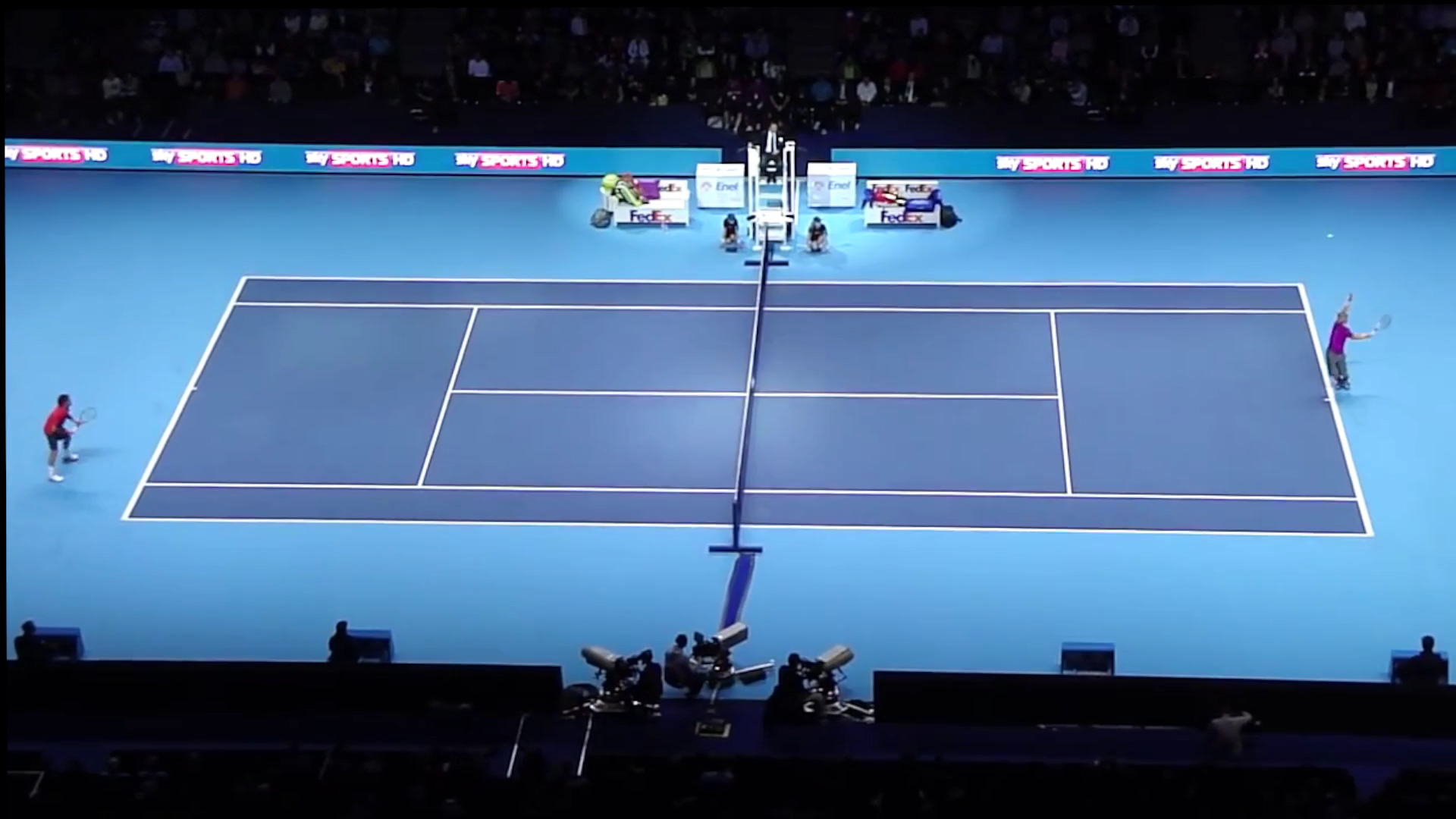} &
\includegraphics[height=0.2\textwidth]{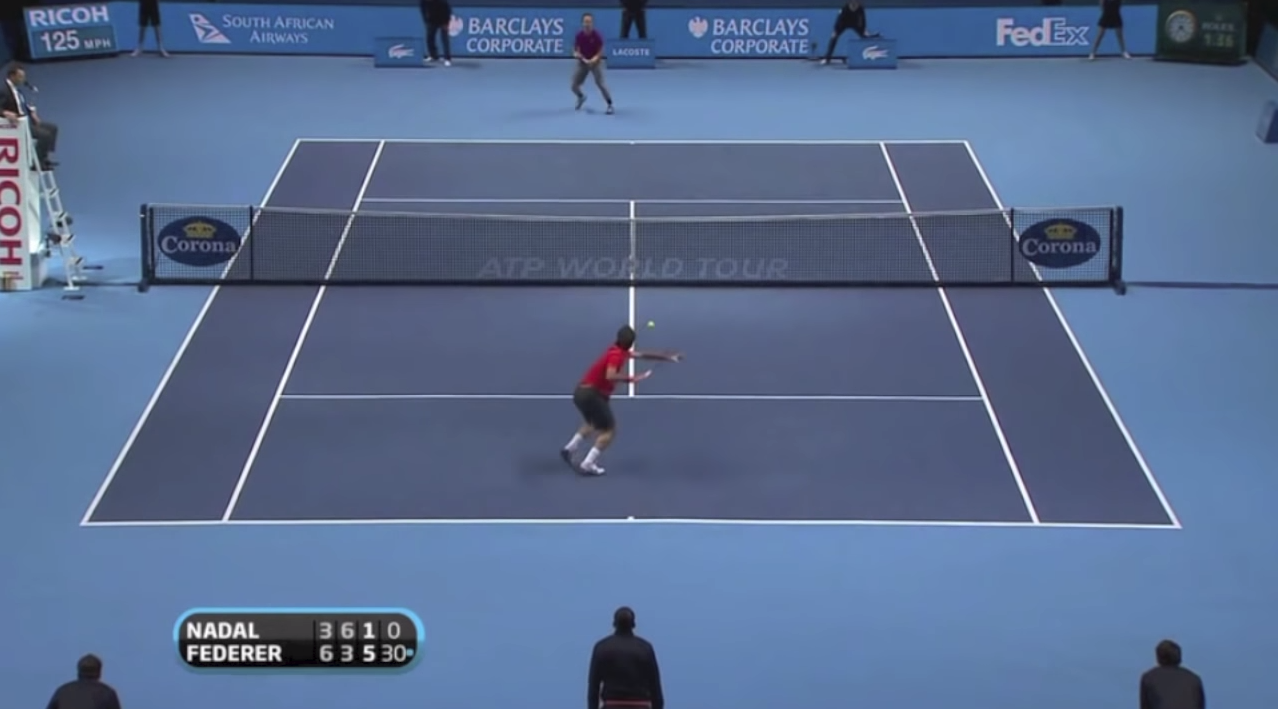} &
\includegraphics[height=0.2\textwidth]{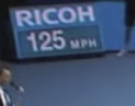} \\ 
Spectator's view & Front view (speed in top left) &  Cropped \\
\end{tabular}
\caption[Radar gun measurements]{Radar gun measurements. Speed was automatically estimated by TbD-NC method from the video on the left. Ground truth acquisition from YouTube video is shown in the middle and the right images. Table~\ref{tbl:atp} compares estimates to the ground truth.}
\label{fig:atp_imgs}
\end{figure}

\paragraph{Speed estimation compared to radar guns.}
In sports, such as tennis, radar guns are commonly used to estimate the speed of serves. In this case, only the maximum speed is measured and the strongest signal usually happens immediately after the racquet hits the ball. 
Hrabalík~\cite{hrabalik} gathered the last 10 serves of the final match of 2010 ATP  World Tour. The serves were found on YouTube from a spectator's viewpoint. Ground truth was available from another footage which showed the measured speeds from radar guns (example in Fig.~\ref{fig:atp_imgs}). A real-time version of FMO detector in~\cite{hrabalik} achieved precise estimates of the speeds with the average error of 4.7 \%, where the error is computed as an absolute difference to the ground truth velocity divided by the ground truth velocity. 

Unfortunately, the ATP footage from spectator's viewpoint is of a very poor quality and the tennis ball is visible only as several pixels. Deblurring does not perform well when a video has low resolution or the object of interest is poorly visible. To test only the performance of full trajectory estimation (TbD-NC), we manually simulated FMO detector by annotating only start and end points of the ball trajectory in several frames after the hit for every serve. Then the time-stamp $t_{hit}$ is found, such that the final trajectory $\C_f(t_{hit})$ at this point is the closest to the hit point. Then $\| \C'_f(t_{hit}) \|$ is the speed measured by TbD-NC.
The pixel-to-miles transformation was computed by measuring the court size in the video (1519 pixels) and dividing it by the tennis standards (78~feet). The camera frame rate was set to the standard 29.97 fps. Additionally, due to severe camera motion, the video was stabilized by computing an affine transformation between consecutive frames using feature matching as in~\cite{fmo}. 
Table~\ref{tbl:atp} compares the speed estimated by TBD-NC and FMO methods to the ground truth from the radar. The proposed TbD-NC method is more precise than the FMO method and in several cases the speed is estimated with GT error close to zero. 

\begin{figure}[t]
\centering
\includegraphics[width=0.7\textwidth]{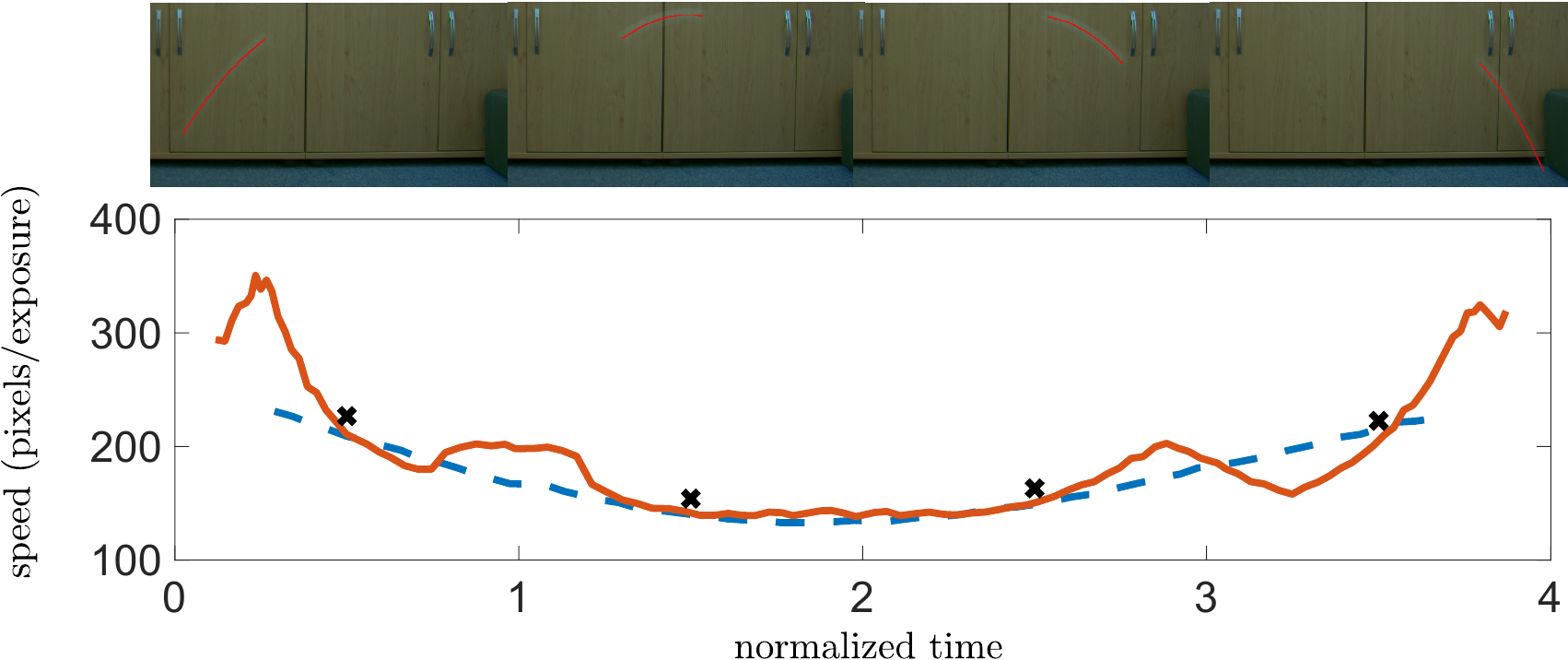}
\caption{Estimating the object velocity from blur kernels. In four consecutive frames (top row), object trajectories were estimated with TbD. The bottom plot shows the velocity calculated from the blur kernels (solid red) and the ground-truth (dashed blue line) obtained by a high-speed camera. Black crosses show the average velocity per frame calculated from the trajectory length.}
\label{fig:speed_blur}
\end{figure}

\paragraph{Speed from the blur kernel.}
Apart from estimating speed by taking the norm of the derivative of $\C_f(t)$, we can also directly estimate speed from the blur kernel $H$. The values in the blur kernel are directly proportional to time the object spent in that location. For example, if half of the exposure time the object was moving with a constant velocity and than it stopped and stayed still, the blur kernel will have constant intensity values terminated with a bright spot that will be equal to the sum of intensities of all other pixels. Estimating speed from blur intensity values is however not very reliable due to noise in $H$. 
Fig.~\ref{fig:speed_blur} illustrates a case where this approach works. All pixels in the blur kernel $H$ which lay on the trajectory $\C$ are used for calculating the object velocity.

\begin{table}[t]
\centering
\setlength{\tabcolsep}{1.0mm}
\caption{Estimation of radius, speed and gravity by TbD-NC on the TbD dataset~\cite{tbd}. 
The speed estimation is compared to GT from a high-speed camera. Radius is calculated when assuming Earth gravity, or vice versa. Standard object sizes are taken as GT for radius. 
}
\begin{center}
\begin{tabular}{l|c|c|c|c|c|c}
\hline
\multirow{2}{*}{Sequence} & \multicolumn{1}{c|}{Speed  } & \multicolumn{3}{c|}{Radius} & \multicolumn{2}{c}{Gravity} \\ \cline{2-7} 
 & Mean Diff. [$r / \epsilon$] & GT [cm] & Est. [cm] & Err. [\%]  & Est. [$m s^{-2}$] & Err. [\%]  \\ 

 \hline

badminton\_white & 0.57  & - & - & - & - & -    \\  
badminton\_yellow & 0.65  & - & - & - & - & -    \\  
pingpong & 0.66  & 2.00  & 1.99  & 0.3  & 9.53  & 2.8   \\  
tennis & 0.56  & - & - & - & - & -    \\  
volleyball & 0.45  & 10.65  & 10.47  & 1.7  & 10.50   & 7.2   \\  
throw\_floor & 0.61  & 3.60  & 3.47  & 3.7  & 10.21   & 4.2   \\  
throw\_soft & 0.42  & 3.60  & 3.72  & 3.3  & 9.52   & 2.9    \\  
throw\_tennis & 1.31  & 3.43  & 3.69  & 7.6  & 9.19   & 6.2  \\  
roll\_golf & 2.54  & - & - & - & - & -    \\  
fall\_cube & 2.24  & 2.86  & 2.63  & 8.0  & 10.66   & 8.8  \\  
hit\_tennis & 0.43  & - & - & - & - & -    \\  
hit\_tennis2 & 1.28  & - & - & - & - & -    \\  
\hline 
Average & 0.98 & - & - & 4.1  & 9.93  & 5.3     \\

\hline
\end{tabular}
\end{center}
\label{tbl:est}
\end{table}



\paragraph{Shape and gravity estimation.}
In many situations, gravity is the only force that has non-negligible influence.
Then, fitting polynomials of second order is sufficient.
If parameters of the polynomial are estimated correctly, and the real gravity is given, then transforming pixels to meters in the region of motion is feasible. Gravity is represented by a parameter $a$, which has units $[\text{px} (\frac{1}{f} s)^{-2}]$, where the frame rate is denoted by $f$. If we assume the gravity of Earth $g \approx 9.8 [m s^{-2}]$, $f$ is known and $a$ is estimated by curve fitting, the formula to convert pixels to meters becomes $p = g / (2 a f^2)$, 
where $p$ are meters in one pixel on the object in motion. The radius estimation by this approach is shown in Table~\ref{tbl:est}. Only half of the TbD dataset is used, \ie sequences where the object was undergoing only motion given by the gravity (throw, fall, ping pong, volleyball). In other cases such as roll and hit, the gravity has almost no influence and this approach cannot be used. The badminton sequences have large air resistance and the tennis sequence was recorded outside during strong wind. When gravity was indeed the only strong force, the estimation has average error 4.1 \%. 
The variation of gravity on Earth is mostly neglectable, but knowing exact location where videos have been recorded might even improve results.
Alternatively, when the real object size is known, we can estimate gravity, \eg when throwing objects on another planet and trying to guess which planet it is. 
In this case, the formula can be rewritten to estimate $g$. Results are also shown in Table~\ref{tbl:est} and the average error is 5.3 \% when compared to the gravity on Earth. This shows robustness of the approach in both estimating radius and gravity.

\paragraph{Temporal super-resolution.}
Among other applications of TbD-NC are fast moving object removal and temporal super-resolution. The task of temporal super-resolution stands for creating a high-speed camera footage out of a standard video and consists of three steps. First, a video free of fast moving objects is produced which is called fast moving object removal. For all FMOs which are found in every frame, we replace them with the estimated background. Second, intermediate frames between adjacent frames are calculated as their linear interpolation. Objects which are not FMOs will look natural after linear interpolation. Then, trajectory $\C_f(t)$ is split into the required number of pieces, optionally with shortening to account for the desired exposure fraction. Third, the object model $(F, M)$ is estimated and used to synthesize the formation model with FMOs~\eqref{eq:acquisition_model}. Examples of these applications are provided in the supplementary material.

\section{Conclusion}
\label{sec:conc}
We proposed a non-causal Tracking by Deblatting (TbD-NC) which estimates accurate and complete trajectories of fast moving objects in videos. TbD-NC is based on globally minimizing an optimality condition which is done by dynamic programming. High-order polynomials are then fitted to trajectory segments without bounces. The method performs well on the recently proposed TbD dataset and complete failures appear 10 times less often. From the estimated trajectories, we are able to calculate precise object properties such as velocity or shape. The speed estimation is compared to the data obtained from a high-speed camera and radar guns. Novel applications such as fast moving objects removal and temporal super-resolution are shown.

\blfootnote{\textbf{Acknowledgements.} This work was supported by the Czech Science Foundation grant GA18-05360S and the Czech Technical University student grant SGS17/185/OHK3/3T/13.}

\bibliographystyle{splncs04}
\bibliography{main}

\end{document}